\documentclass{article}
\usepackage[final]{neurips_2019}
\usepackage[utf8]{inputenc}
\usepackage{microtype}
\usepackage{graphicx}
\usepackage{subfigure}
\usepackage{booktabs}
\usepackage{url}

\usepackage{hyperref}

\usepackage{mathtools}
\usepackage{amsthm}
\usepackage{amsmath}
\usepackage[psamsfonts]{amssymb}
\usepackage[mathscr]{euscript}
\usepackage{bbm}
\usepackage{multirow}

\usepackage{MnSymbol}                                                                  \usepackage{xpatch}

\hypersetup{
  colorlinks   = true, 
  urlcolor     = blue, 
  linkcolor    = blue, 
  citecolor   = blue 
}

\newcommand{\ignore}[1]{}

\title{Can You Really Backdoor Federated Learning?}
\author{Ziteng Sun\thanks{Work done while ZS was an intern at Google.}\\
  Cornell University \\
  \texttt{zs335@cornell.edu} \\
  \And
  Peter Kairouz \\
  Google\\
  \texttt{kairouz@google.com} \\
  \And 
  Ananda Theertha Suresh\\
  Google\\
  \texttt{theertha@google.com} \\
  \And
  H. Brendan McMahan\\
  Google\\
  \texttt{mcmahan@google.com}\\}

\usepackage{natbib}
\usepackage{graphicx, amsmath, amsfonts}
\setcitestyle{numbers}

\begin{document}

\maketitle

\begin{abstract}
The decentralized nature of federated learning makes detecting and defending against adversarial attacks a challenging task. This paper focuses on backdoor attacks in the federated learning setting, where the goal of the adversary is to reduce the performance of the model on targeted tasks while maintaining a good performance on the main task. Unlike existing works, we allow non-malicious clients to have correctly labeled samples from the targeted tasks. We conduct a comprehensive study of backdoor attacks and defenses for the EMNIST dataset, a real-life, user-partitioned, and non-iid dataset. We observe that in the absence of defenses, the performance of the attack largely depends on the fraction of adversaries present and the ``complexity'' of the targeted task. Moreover, we show that norm clipping and ``weak'' differential privacy mitigate the attacks without hurting the overall performance. We have implemented the attacks and defenses in TensorFlow Federated (TFF), a TensorFlow framework for federated learning. In open sourcing our code, our goal is to encourage researchers to contribute new attacks and defenses and evaluate them on standard federated datasets.
\end{abstract}

\section{Introduction}
Modern machine learning systems can be vulnerable to various kinds of failures, such as bugs in preprocessing pipelines and noisy training labels, as well as attacks that target each step of the system's training and deployment pipelines. Examples of attacks include data and model update poisoning~\cite{Biggio:2012:PAA:3042573.3042761, DBLP:conf/ndss/LiuMALZW018}, model evasion~\cite{DBLP:journals/corr/SzegedyZSBEGF13,Biggio:2012:PAA:3042573.3042761,DBLP:journals/corr/GoodfellowSS14}, model stealing~\cite{DBLP:conf/uss/TramerZJRR16}, and data inference attacks on users' private training data~\cite{DBLP:conf/sp/ShokriSSS17}.

The distributed nature of federated learning~\cite{mcmahan17fedavg}, particularly when augmented with secure aggregation protocols~\cite{bonawitz17secagg}, makes detecting and correcting for these failures and attacks a particularly challenging task. Adversarial attacks can be broadly classified into two types based on the goal of the attack, untargeted or targeted attacks. Under untargeted attacks~\cite{NIPS2017_6617, pmlr-v80-mhamdi18a, damaskinos2018asynchronous}, the goal of the adversary is to corrupt the model in such a way that it does not achieve a near-optimal performance on the main task at hand (e.g., classification) often referred to as the primary task. Under targeted attacks (often referred to as backdoor attacks) \cite{chen2017targeted, liao2018backdoor, gu2019badnets}, the goal of the adversary is to ensure that the learned model behaves differently on certain targeted sub-tasks while maintaining good overall performance on the primary task. For example, in image classification, the attacker may want the model to misclassify some ``green cars'' as birds while ensuring that other cars are correctly classified. 

For both targeted and untargeted attacks, the attacks can be further classified into two types based on the capability of the attacker, \emph{model update poisoning} or \emph{data poisoning}. In data poisoning attacks~\cite{Biggio:2012:PAA:3042573.3042761, steinhardt2017certified, Xiao:2015:SVM:2779626.2779777, Mei:2015:UMT:2886521.2886721, huber1997robustness}, the attacker can change a subset of all the training samples which is unknown to the learner. In federated learning systems, since the training process is done on local devices, fully compromised clients can change the model update completely, which is called a model poisoning attack~\cite{bagdasaryan2018backdoor, pmlr-v97-bhagoji19a}. Model update poisoning attacks are even harder to counter when secure aggregation (SecAgg)~\cite{bonawitz17secagg}, which ensures that the server cannot inspect each user's update, is deployed in the aggregation of local updates.

Since untargeted attacks reduce the overall performance of the primary task, they are easier to detect. On the other hand, backdoor targeted attacks are harder to detect as the goal of the adversary is often unknown a priori. Hence, following~\cite{bagdasaryan2018backdoor, pmlr-v97-bhagoji19a}, we consider targeted model update poisoning attacks and refer to them as backdoor attacks. Existing approaches against backdoor attacks~\cite{steinhardt2017certified, liu2018fine, tran2018spectral, pmlr-v97-diakonikolas19a, wang2019neural, pmlr-v97-shen19e} either require a careful examination of the training data or full control of the training process at the server, which may not apply in the federated learning case. We evaluate various attacks proposed in recent papers and defenses on a medium scale federated learning task with more realistic parameters using TensorFlow Federated~\cite{web:TFF}. Our goal, in open sourcing our code, is to encourage researchers to evaluate new attacks and defenses on standard tasks.

\section{Backdoor Attack Scenario}
\label{sec:backdoor_scenario}
We consider the notations and definitions of federated learning as defined in \cite{mcmahan17fedavg}.\footnote{While \cite{mcmahan17fedavg} considers relatively small problems, in more realistic scenarios for mobile devices we might have $K = 10^7$ or higher, with the number of clients selected $C \cdot K$ typically constant, say 100 to 1000 per round.} In particular, let $K$ be the total number of users. At each round $t$, the server randomly selects $ C \cdot K$ clients for some $C < 1$. Let $S_t$ be this set and $n_k$ be the number of samples at client $k$. Denote the model parameters at round $t$ by $w_t$.
Each selected user computes a model update, denoted by $\Delta w^k_t$, based on their local data. 
The server updates its model by aggregating the $\Delta w^k_t$'s, i.e.,
\[
w_{t+1} = w_t + \eta \frac{\sum_{k \in S_t} n_k \Delta w^k_{t}}{\sum_{k \in S_t} n_k}.
\]
where $\eta$ is the server learning rate.
We model the parameters of backdoor attacks as follows.

\textbf{Sampling of adversaries.} If $\epsilon$ fraction of the clients are completely compromised, then each round may contain anywhere between $0$ and $\min(\epsilon \cdot K, C\cdot K)$ adversaries. Under random sampling of clients, the number of adversaries in each round follows a hypergeometric distribution. We refer to this attack model as the \emph{random sampling} attack. Another model we consider in this work is the \emph{fixed frequency} attack, where a single adversary appears in every $f$ rounds~\cite{bagdasaryan2018backdoor, pmlr-v97-bhagoji19a}. For a fair comparison between the two attack models, we set the frequency to be inversely proportional to the number of total number of attackers (i.e.,  $f = 1/(\epsilon \cdot C \cdot K)$).


\textbf{Backdoor tasks.} Recall that in backdoor attacks, the goal of the adversary is to ensure that the model fails on some targeted tasks. For example, in text classification the backdoor task might be to suggest a particular restaurant's name after observing the phrase \textit{``my favorite restaurant is''}.
Unlike \cite{bagdasaryan2018backdoor, pmlr-v97-bhagoji19a}, we allow non-malicious clients to have correctly labeled samples from the targeted backdoor tasks. For instance, if the adversary wants the model to misclassify some green cars as birds, we allow non-malicious clients to have samples from these targeted green cars correctly labeled as cars. 

Further, we form the backdoor task by grouping examples from multiple selected ``target clients''. Since examples from different target clients follow different distributions, we refer to the number of target clients as the ``number of backdoor tasks'' and explore its effect on the attack's success rate. Intuitively, the more backdoor tasks we have, the richer the feature space the attacker is trying to break, and therefore the harder it is for the attacker to successfully backdoor the model without breaking its performance on the main task. 

\section{Model Update Poisoning Attacks}
We focus on model update poisoning attacks based on the model replacement paradigm proposed by~\cite{bagdasaryan2018backdoor, pmlr-v97-bhagoji19a}. When only one attacker is selected in round $t$ (WLOG assume it is client 1), the attacker attempts to replace the whole model by a backdoored model $w^*$ by sending
\begin{align} \label{eqn:update_boost}
    \Delta w^1_{t} =  \beta(w^* - w_t).
\end{align}
where $\beta = \frac{\sum_{k\in S_t}n_k}{\eta n_k}$ is a boost factor. Then we have \[
    \Delta w_{t+1} =  w^* + \eta \frac{\sum_{k \in S_t, k \neq 1} n_k \Delta w^k_{t}}{\sum_{k \in S_t} n_k},
    \]
which will be in a small neighbourhood of $w^*$ if we assume the model has sufficiently converged and hence the other updates $\Delta w^k_t$ for $k > 1$ are small. If multiple attackers appear in the same round, we assume that they can coordinate with each other and divide this update evenly.

\textbf{Obtaining a backdoored model.} To obtain a backdoored model $w^*$, we assume that the attacker has a set $D_{\text{mal}}$ which describes the backdoor task -- for example, different kinds of green cars labeled as birds. We also assume the attacker has a set of training samples generated from the true distribution $D_{\text{trn}}$. Note that for practical applications, such data may be harder for the attacker to obtain.

\textbf{Unconstrained boosted backdoor attack. } In this case, the adversary trains a model $w^*$ based on $w_t, D_{\text{mal}}$ and $D_{\text{trn}}$ without any constraints and sends the update based on~\eqref{eqn:update_boost} back to the service provider. One popular training strategy is to initialize with $w_t$ and train the model with $D_{\text{trn}} \cup D_{\text{mal}}$ for a few epoches. This attack generally results in a large update norm and can serve as a baseline.



\textbf{Norm bounded backdoor attack.} Unconstrained backdoor attacks can be defended by norm clipping as discussed below. To overcome this, we consider the norm bounded backdoor attack. Here at each round, the model trains on the backdoor task subject to the constraint that the model update is smaller than $M/\beta$. Thus, model update has norm bounded by $M$ after boosted by a factor of $\beta$. This can be done by training the model using multiple rounds of projected gradient descent, where in each round we train the model using the unconstrainted training strategy and project it back to the $\ell_2$ ball of size $M/\beta$ around $w_t$.

\section{Defenses}

We consider the following defenses for backdoor attacks.

\textbf{Norm thresholding of updates}. Since boosted attacks are likely to produce updates with large norms, a reasonable defense is for the server to simply ignore updates whose norm is above some threshold $M$; in more complex schemes $M$ could even be chosen in randomized fashion. However, in the spirit of investigating what a strong adversary might accomplish, we assume the adversary knows the threshold $M$, and can hence always return malicious updates within this magnitude. Giving this strong advantage to the adversary makes the norm-bounding defense equivalent to the following norm-clipping approach:
\[
\Delta w_{t+1} = \sum_{k \in S_t} \frac{\Delta w^k_{t+1}}{\max( 1, ||\Delta w^k_{t+1}||_2 / M)}.
\]
This model update ensures that the norm of each model update is small and hence less susceptible to the server. 

\textbf{(Weak) differential privacy}. A mathematically rigorous way for defending against backdoor tasks is to train models with differential privacy ~\cite{ma2019data, Dwork06, abadi2016deep}. These approaches were extended to the federated setting by \cite{mcmahan18dplm}, by first clipping updates (as above) and then adding Gaussian noise. We explore the effect of this method. However, traditionally the amount of noise added to obtain reasonable differential privacy is relatively large. Since our goal is not privacy, but instead preventing attacks, we add a small amount of noise that is empirically sufficient to limit the success of attacks. 

\section{Experiments}
In the above backdoor attack framework, we conduct experiments on the EMNIST dataset~\cite{cohen2017emnist, caldas2018leaf}\footnote{Code available at \url{https://github.com/tensorflow/federated/tree/master/tensorflow_federated/python/research/targeted_attack}.}. This dataset is a writer-annotated handwritten digit classification dataset collected from $3383$ users with roughly $100$ images of digits
per user. Each of them has their unique writing style. We train a five-layer convolution neural network with two convolution layers, one max-pooling layer and two dense layers using federated learning in the TensorFlow Federated framework~\cite{web:TFF}. At each round of training, we select $C \cdot K = 30$ clients. Each client trains the model with their own local data for 5 epochs with batch size 20 and client learning rate 0.1. We use a server learning rate of $1$.

In the experiment, we consider the backdoor task as classifying 7s from multiple selected ``target clients'' as 1s. Note that our attack approach does not 
require 7s from other clients to be classified as 1s.
Since 7s coming from different target clients follow different distributions (because they have different writing styles), we refer to the number of target clients as the ``number of backdoor tasks''.

\begin{figure}[tbh!]
\centering
\subfigure[Attack frequency = 1 ($\epsilon = 3.3\%$)\label{fig:freq1}]{\includegraphics[width = 0.49\textwidth] {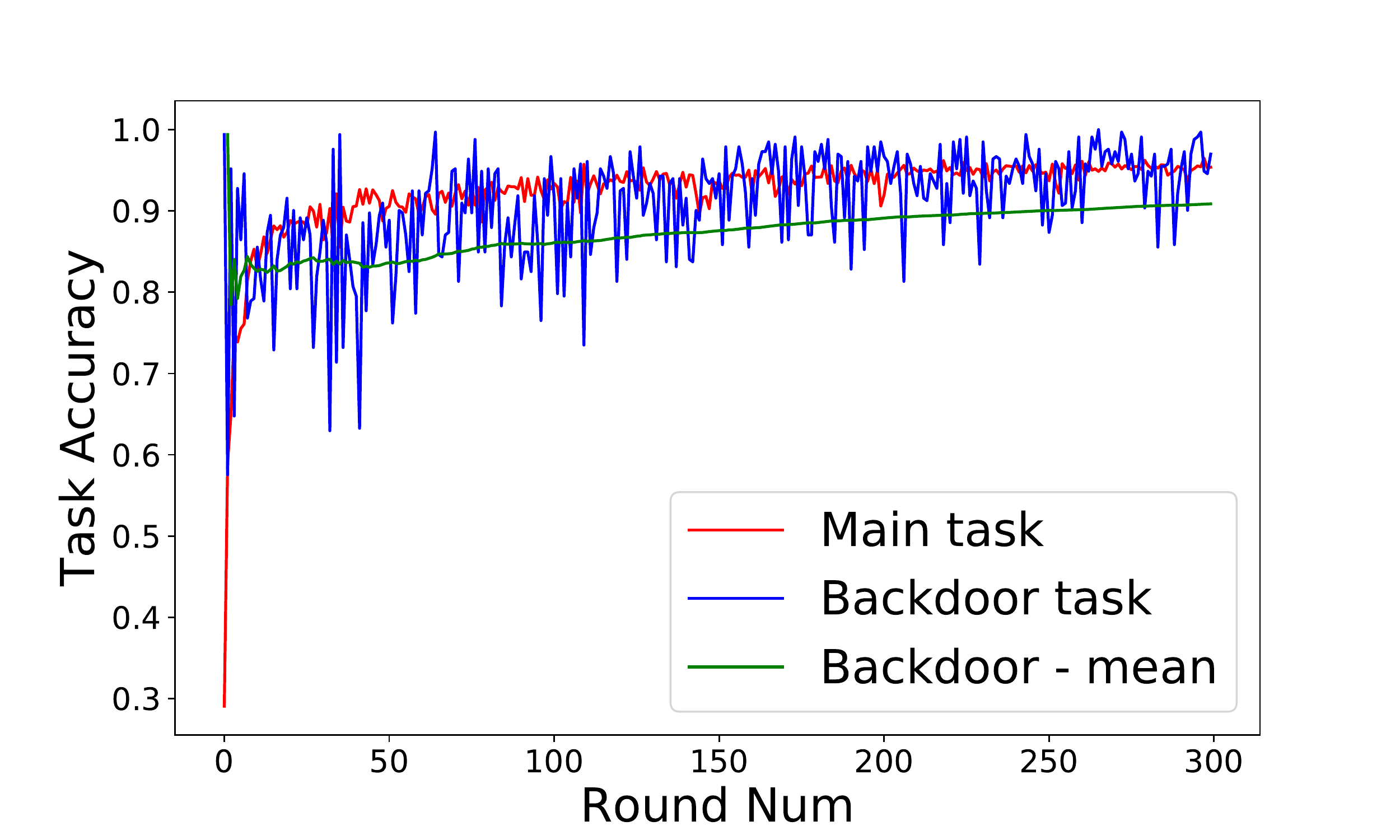}} \hfill \subfigure[Number of attackers = 113 ($\epsilon = 3.3\%$)\label{fig:freq1_random}]{\includegraphics[width = 0.49\textwidth]{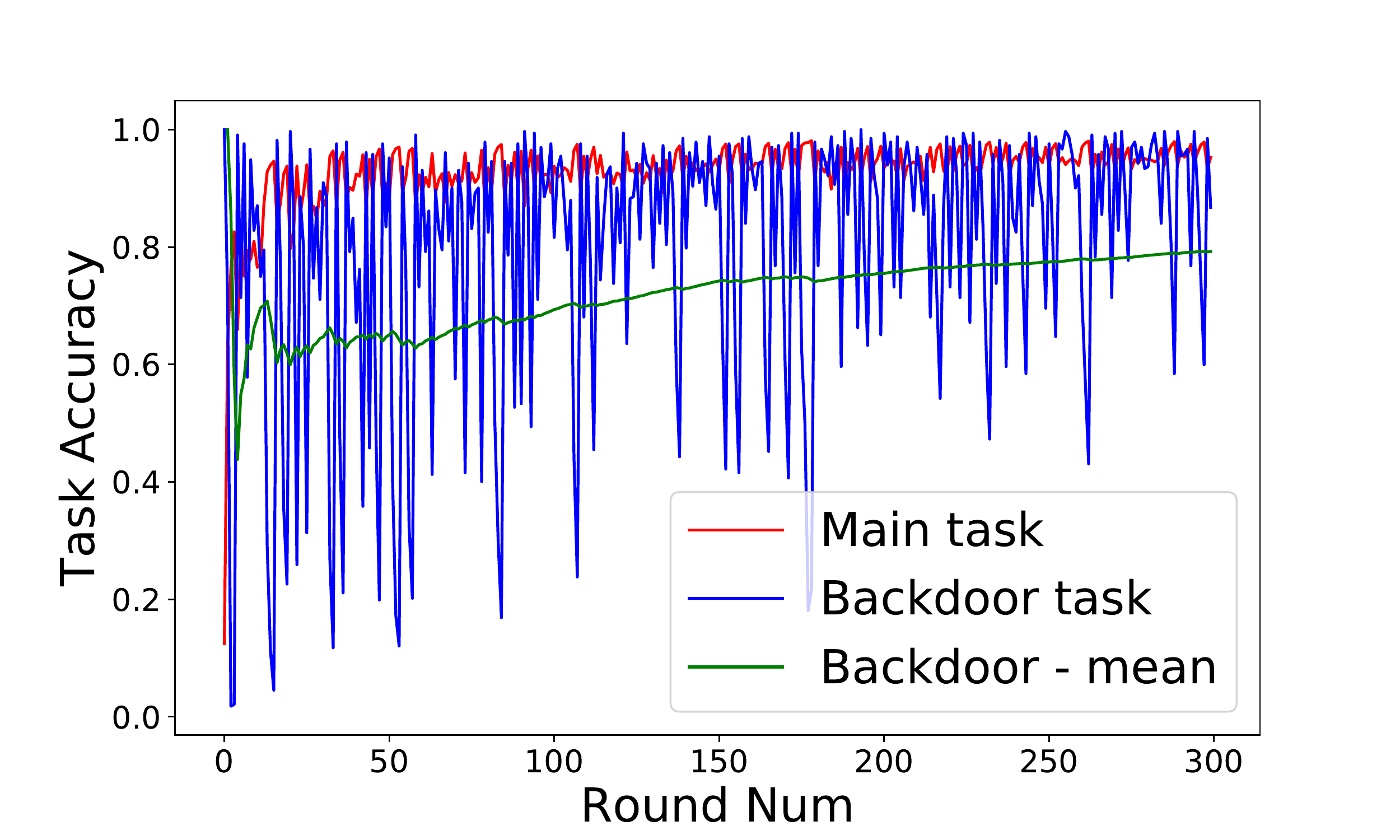}}

\subfigure[Attack frequency = 1/10 ($\epsilon = 0.33\%$)\label{fig:freq10}]{\includegraphics[width = 0.49\textwidth] {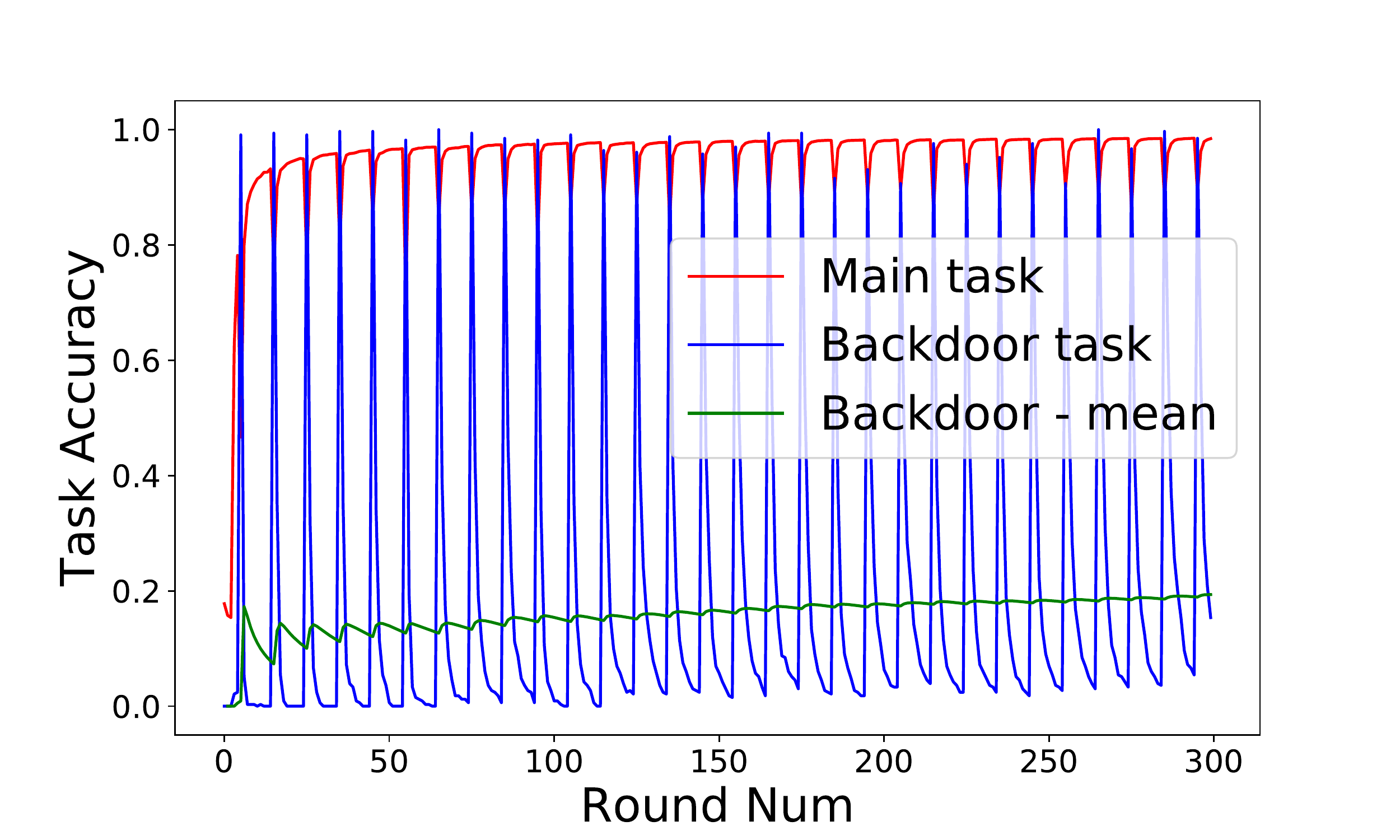}} \hfill 
\subfigure[Number of attackers = 11 ($\epsilon = 0.33\%$)\label{fig:freq10_random}]{\includegraphics[width = 0.49\textwidth]{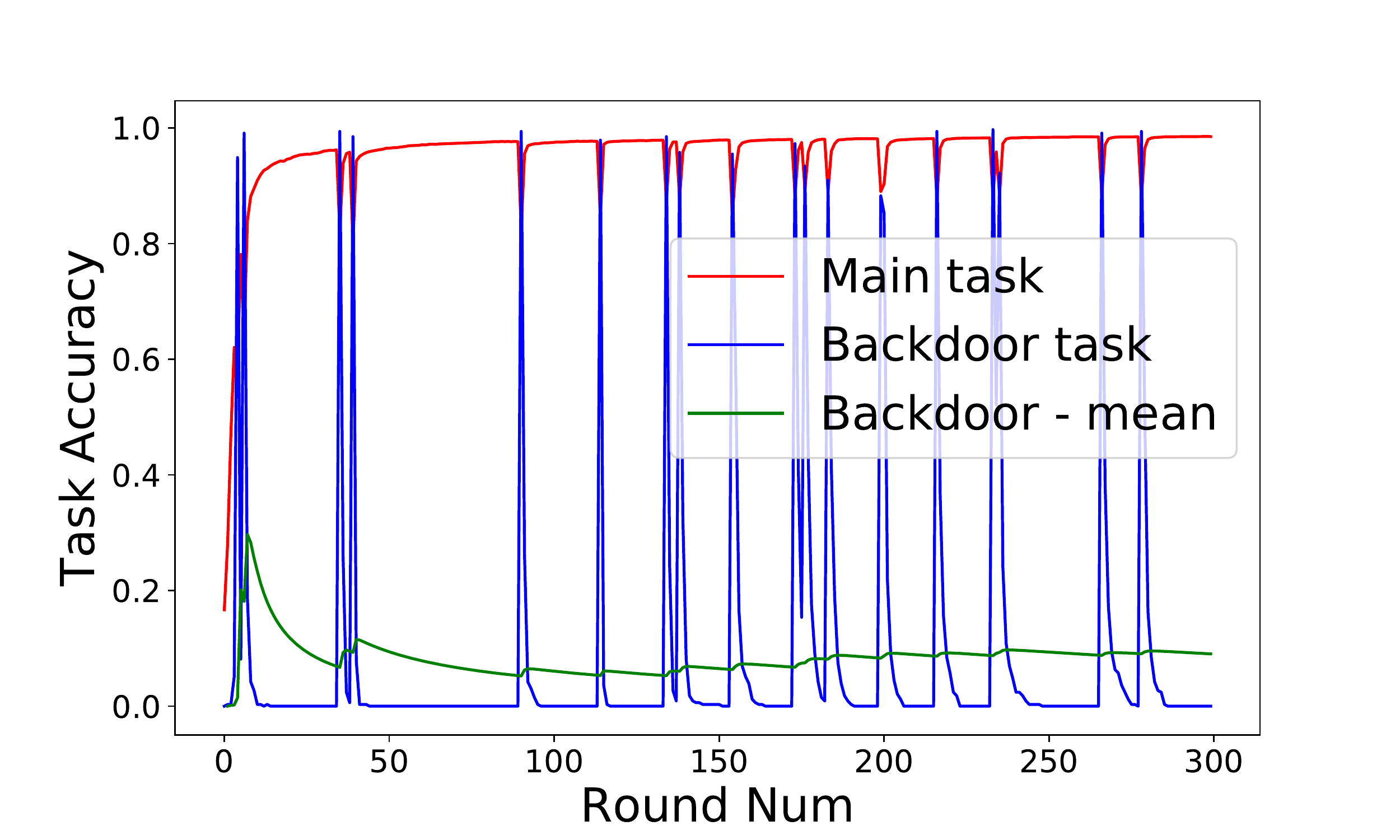}}

\caption{Unconstrained attack for fixed-frequency attacks (left column) and random sampling attack (right column) with different fractions of attackers. Green line is the cumulative mean for the backdoor accuracy.}
\label{fig:unconstrained}
\end{figure}

\begin{figure}[tbh!]
\centering
\subfigure[Attack frequency = 1 ($\epsilon = 3.3\%$)\label{fig:freq1_norm10}]{\includegraphics[width = 0.49\textwidth] {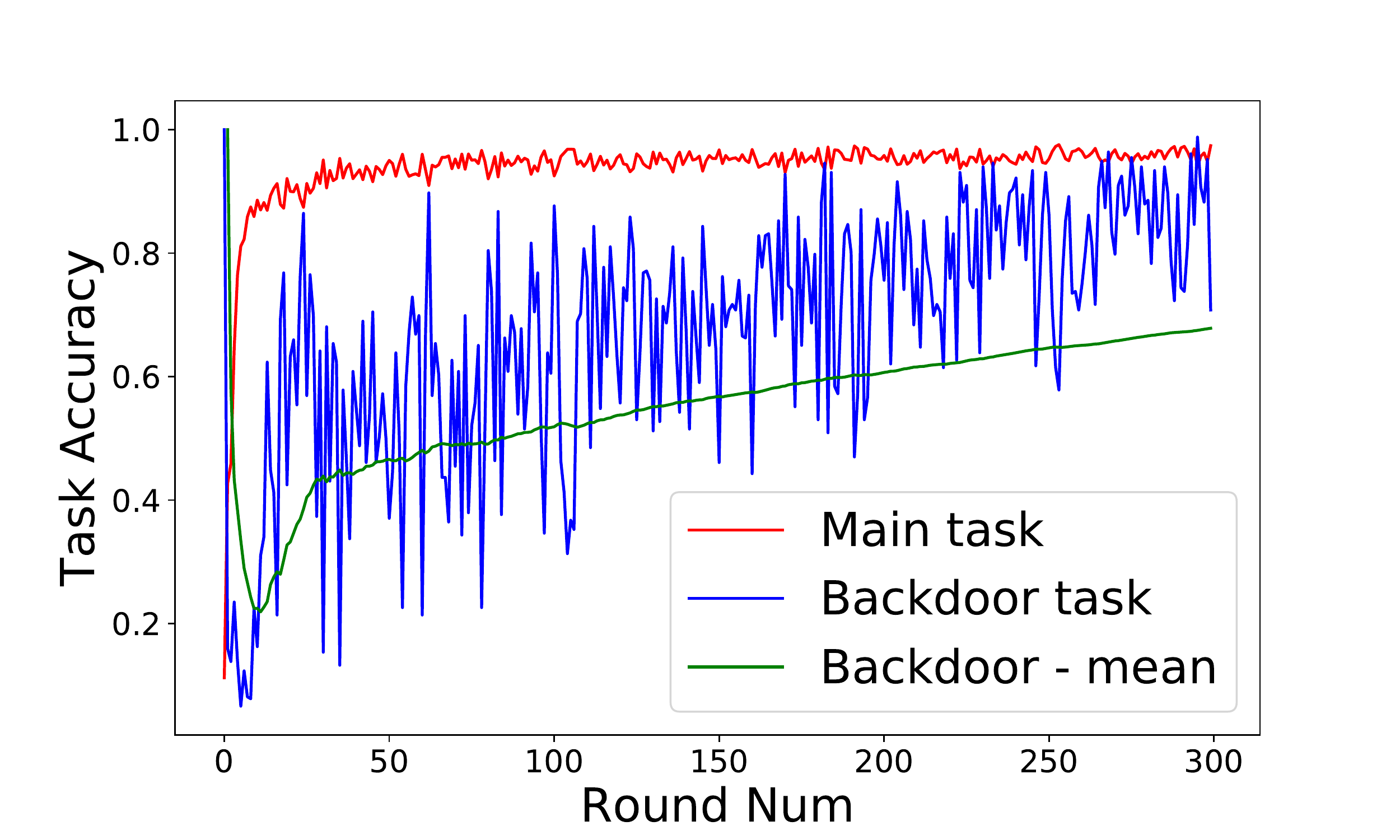}} \hfill 
\subfigure[Number of attackers = 113 ($\epsilon = 3.3\%$)\label{fig:freq1_random_norm10}]{\includegraphics[width = 0.49\textwidth]{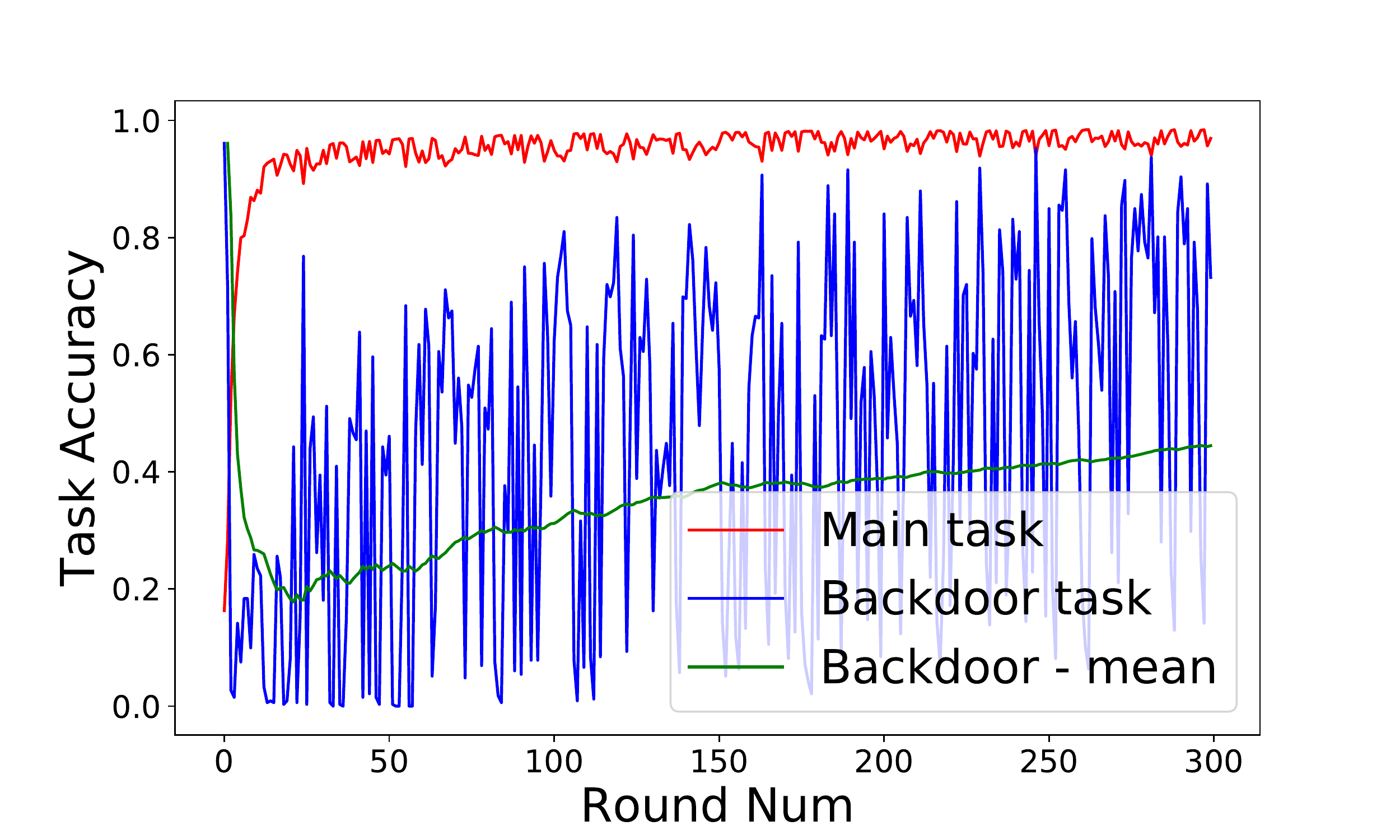}}
\subfigure[Attack frequency = 1/10 ($\epsilon = 0.33\%$)\label{fig:freq10_norm10}]{\includegraphics[width = 0.49\textwidth] {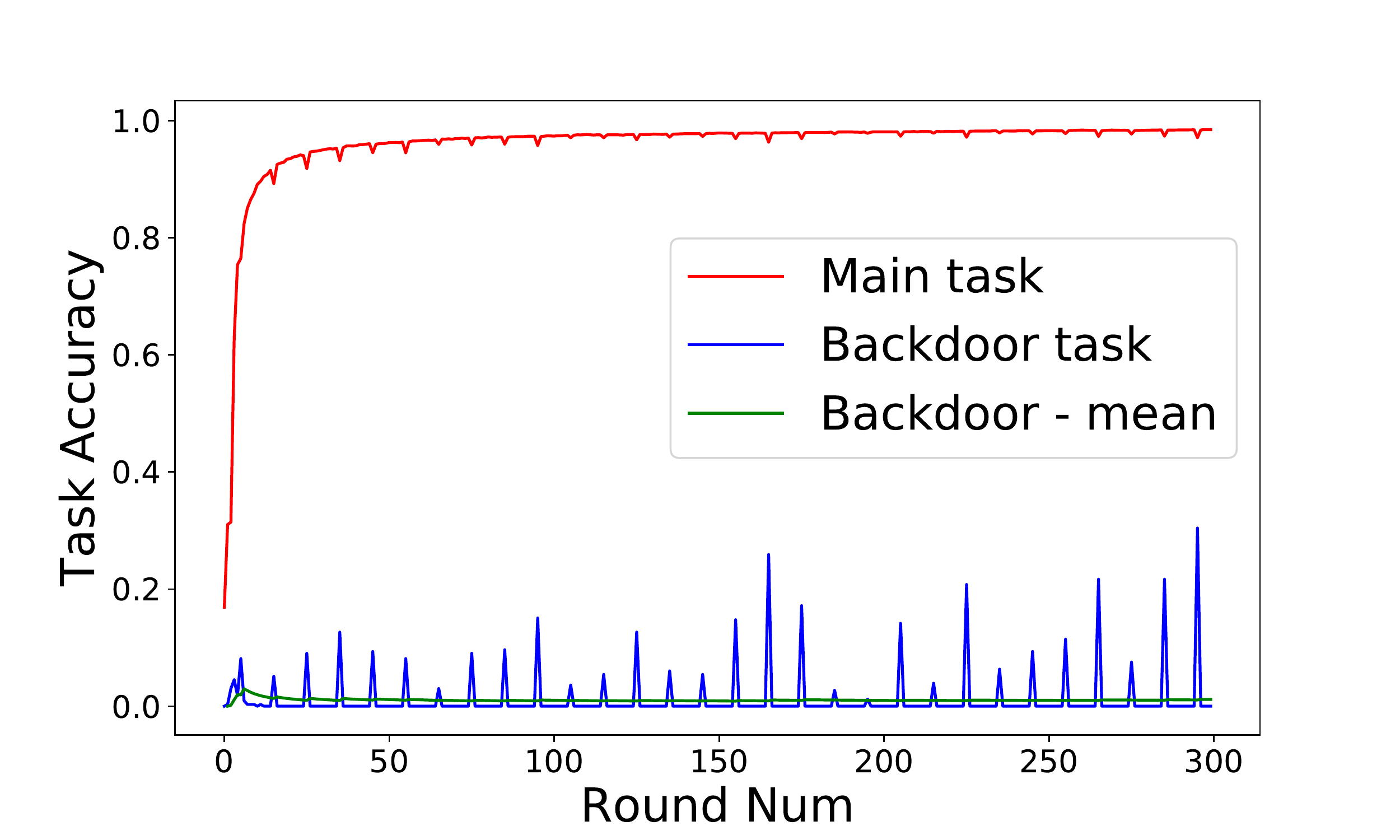}} \hfill 
\subfigure[Number of attackers = 11 ($\epsilon = 0.33\%$)\label{fig:freq10_random_norm10}]{\includegraphics[width = 0.49\textwidth]{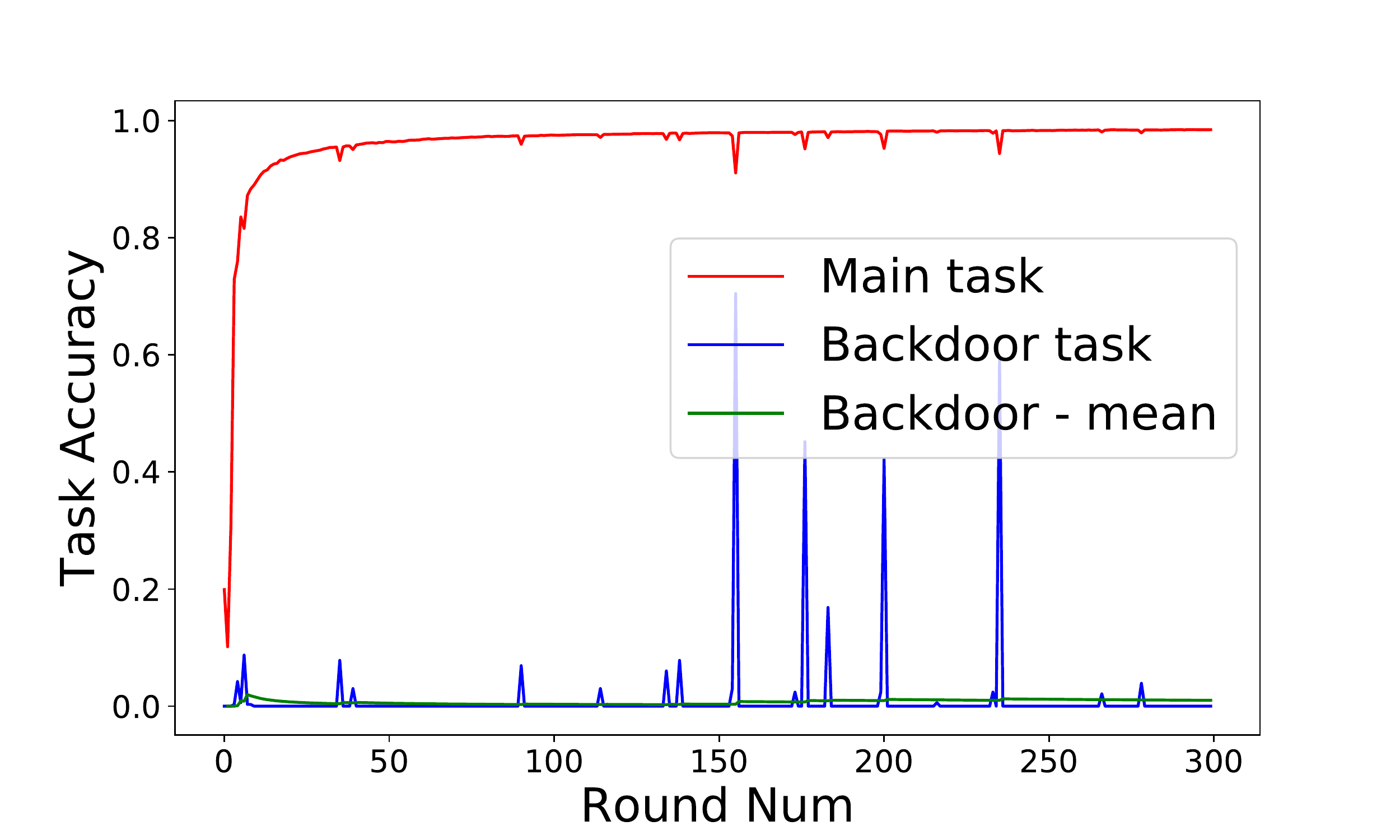}}

\caption{Constrained attack with norm bound 10 for fixed-frequency attacks (left column) and random sampling attack (right column) with different fractions of attackers. Green line is the cumulative mean for the backdoor accuracy.}
\label{fig:constrained}
\end{figure}

\textbf{Random sampling vs. fixed frequency attacks}.  
To begin with, we conduct experiments for the two attack models discussed in Section \ref{sec:backdoor_scenario} under different fractions of adversaries. The results are shown in Figure~\ref{fig:unconstrained} (for unconstrained attack) and Figure~\ref{fig:constrained} (for norm bounded attack). {Additional plots are shown in Figure~\ref{fig:unconstrained_app} and Figure~\ref{fig:constrained_app} in the appendix.} The figures show that both attack models have similar behaviors, despite fixed frequency attacks being slightly more effective than random sampling attacks. Furthermore, in the fixed frequency attack, it is easier to see if the attack happened in a particular round or not. Hence, to provide additional advantage for the attacker and for ease of interpretability, we focus our analysis on fixed-frequency attacks in the rest of this section.

\textbf{Fraction of corrupted users}. In Figure~\ref{fig:unconstrained} and Figure~\ref{fig:constrained} {(also Figure~\ref{fig:unconstrained_app} and Figure~\ref{fig:constrained_app} in the appendix)}, we consider a malicious task with 30 backdoor tasks (around 300 images). We perform unconstrained attacks and norm-bounded attacks with 
$\epsilon = 3.3\%, 0.33\%$
fraction of users being malicious. Both fixed-frequency attack (left column) and random sampling (right column) attacks are considered. For fixed-frequency attack, this corresponds to attacking frequency of $1$ (attacking every round), 
and $1/10$ (once every ten rounds).
From the above experiment, we can infer that the backdoor attack success largely depends on the fraction of adversaries and the performance of backdoor attack degrades as the fraction of fully compromised users falls below $1\%$. 

\begin{figure}[tbh!]
\centering
\subfigure[Backdoor Size = 10\label{fig:backdoor_num_10}]{\includegraphics[width = 0.49\textwidth]{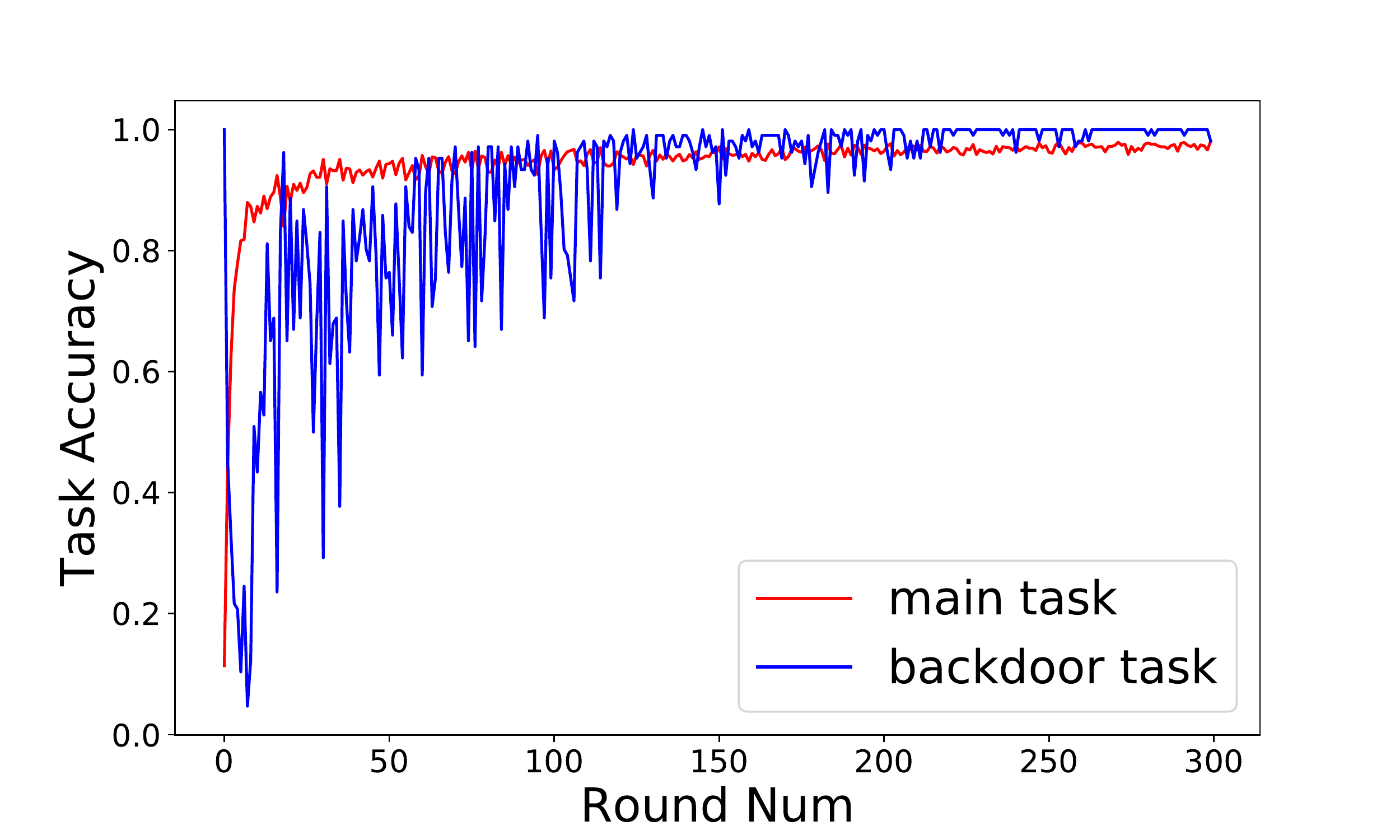}}
\hfill
\subfigure[Backdoor Size = 20\label{fig:backdoor_num_20}]{\includegraphics[width = 0.49\textwidth]{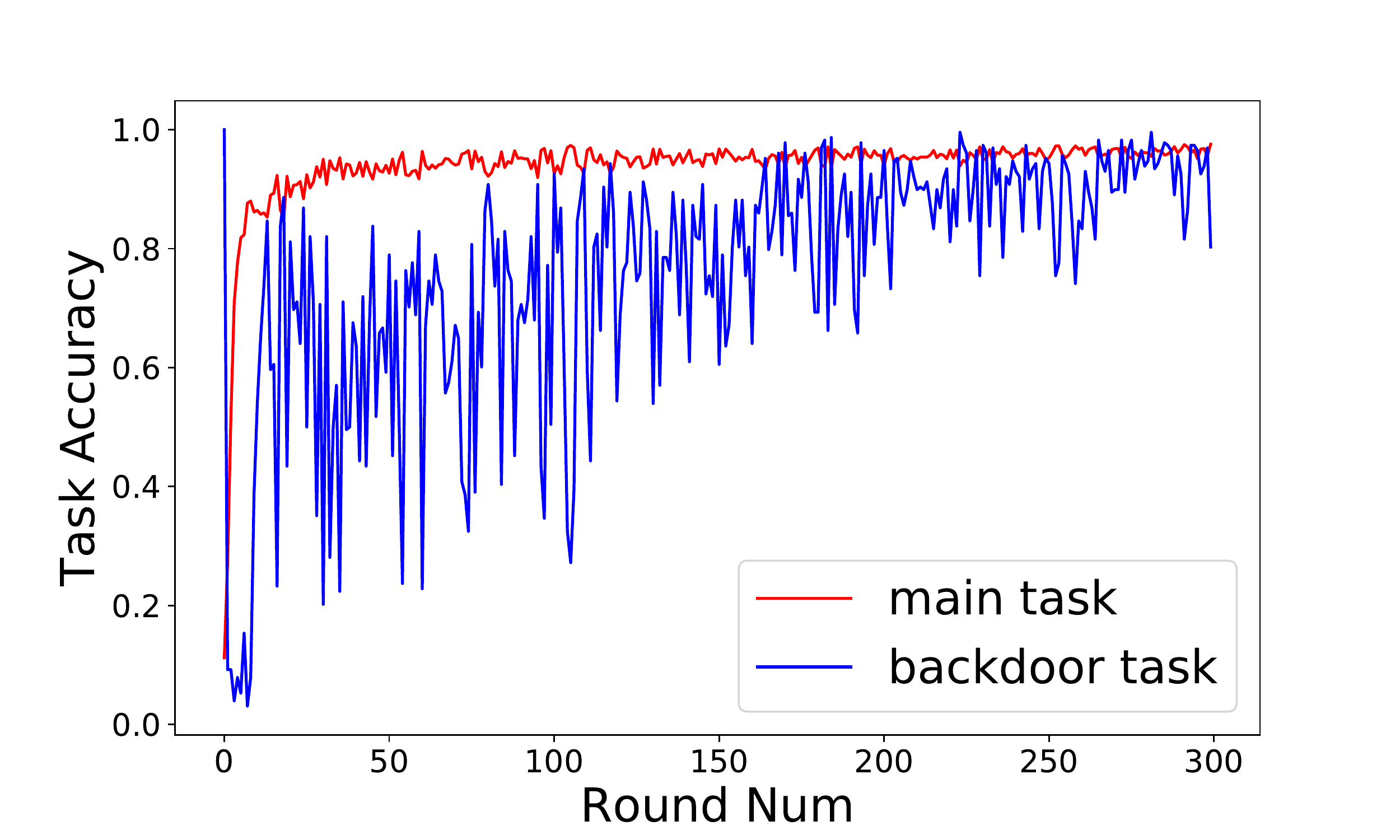}}
\subfigure[Backdoor Size = 30\label{fig:backdoor_num_30}]{\includegraphics[width = 0.49\textwidth]{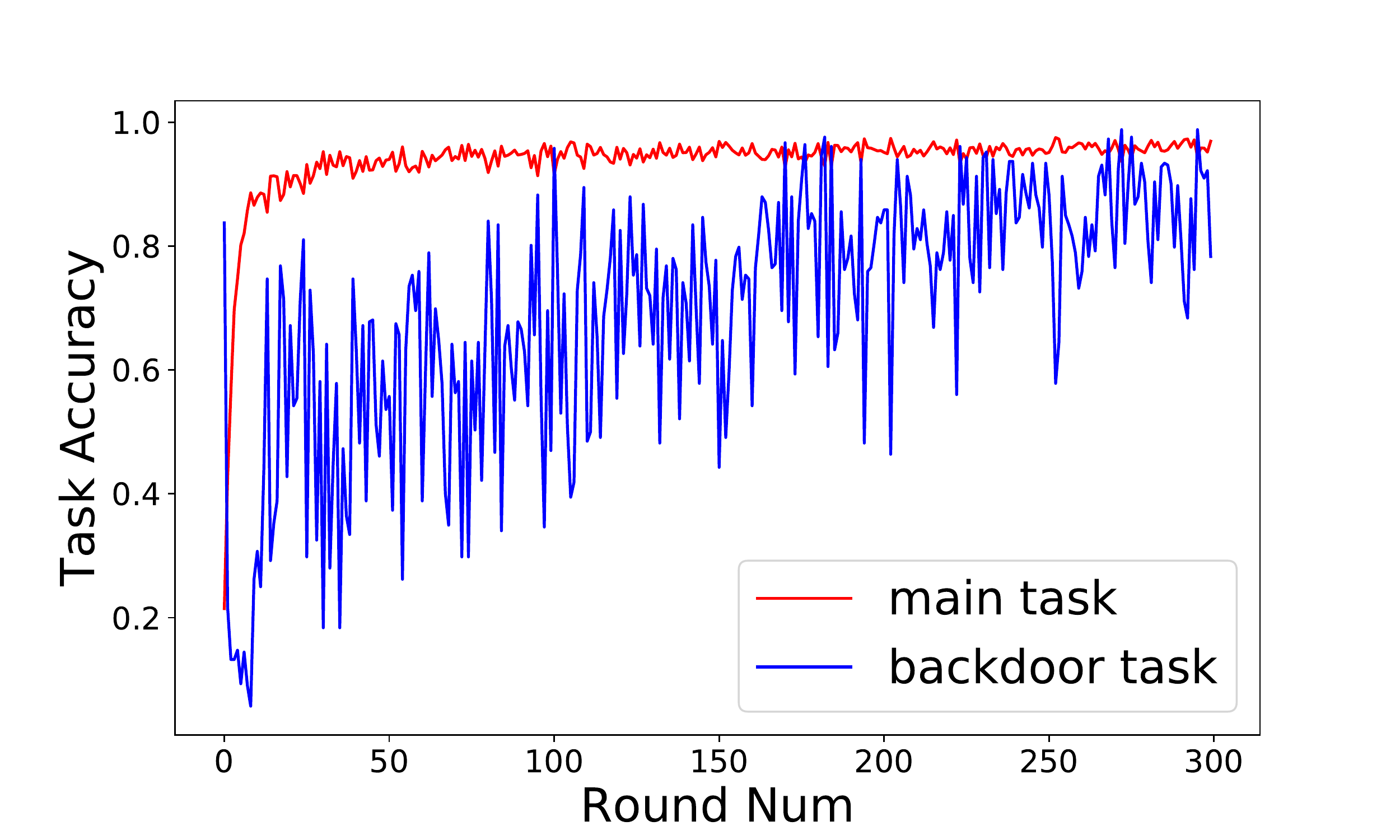}}
\hfill
\subfigure[Backdoor Size = 50\label{fig:backdoor_num_50}]{\includegraphics[width = 0.49\textwidth]{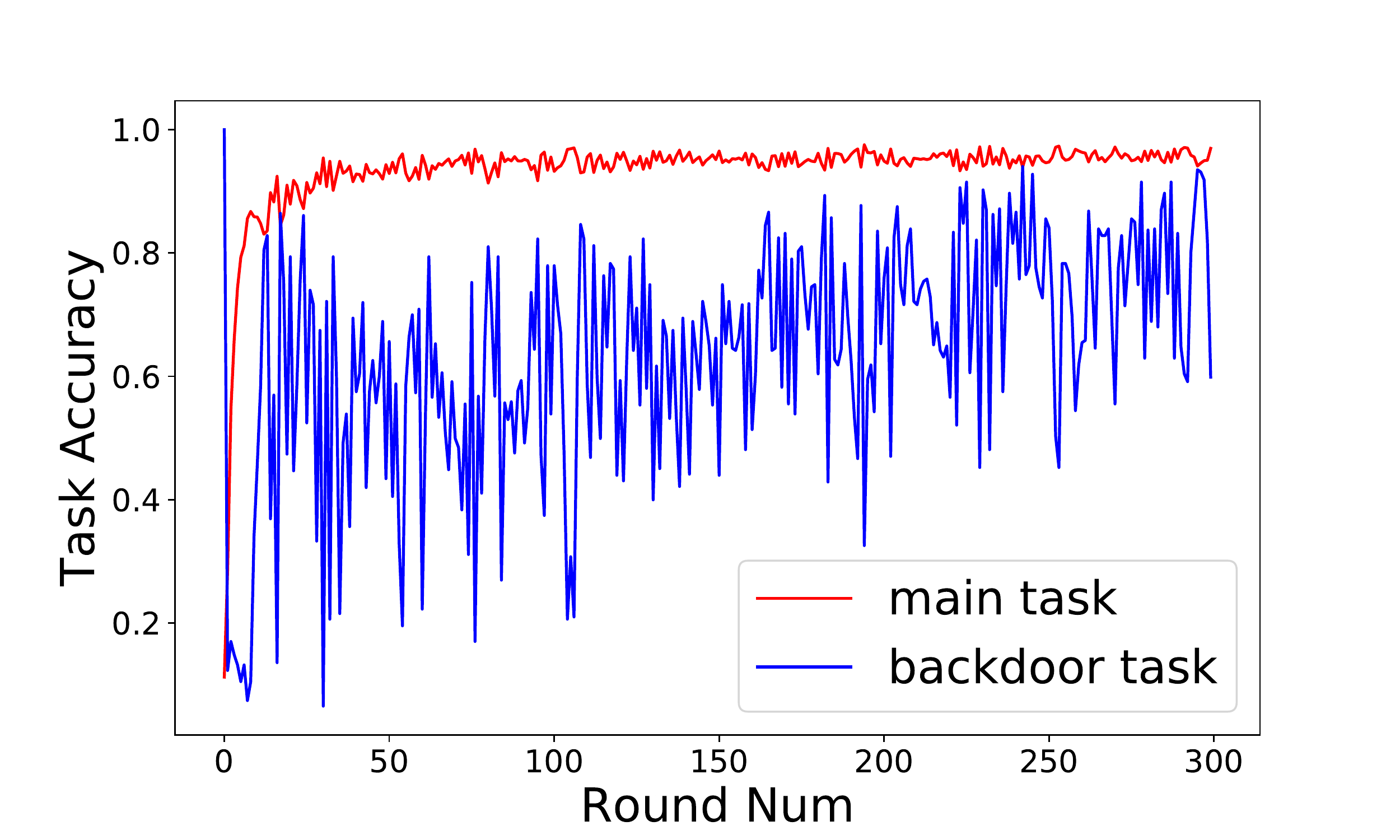}}
\caption{The Effect of Backdoor Size for Constrained Attack with Norm Bound 10.}
\label{fig:backdoor_num}
\end{figure}

\textbf{Number of backdoor tasks}. The number of backdoor tasks affects the performance in two ways. First, the more backdoor tasks we have, the harder it is to backdoor a fixed-capacity model while maintaining its performance on the main task. Second, since we assume benign users have correct samples from the backdoor task, they can correct the attacked model with these samples. In Figure~\ref{fig:backdoor_num}, we consider norm bounded attack with norm bound 10 and 10, 20, 30, 50 backdoor tasks. We can see from the plot that the more backdoor tasks we have, the harder it is to fit a malicious model.

\begin{figure}[tbh!]
\centering
\subfigure[Norm clipping bound: Blue - unattacked baseline, Green: 3, Red: 5, Black: 10]{%
\label{fig:clip} \hfill \includegraphics[width = 0.49\textwidth]{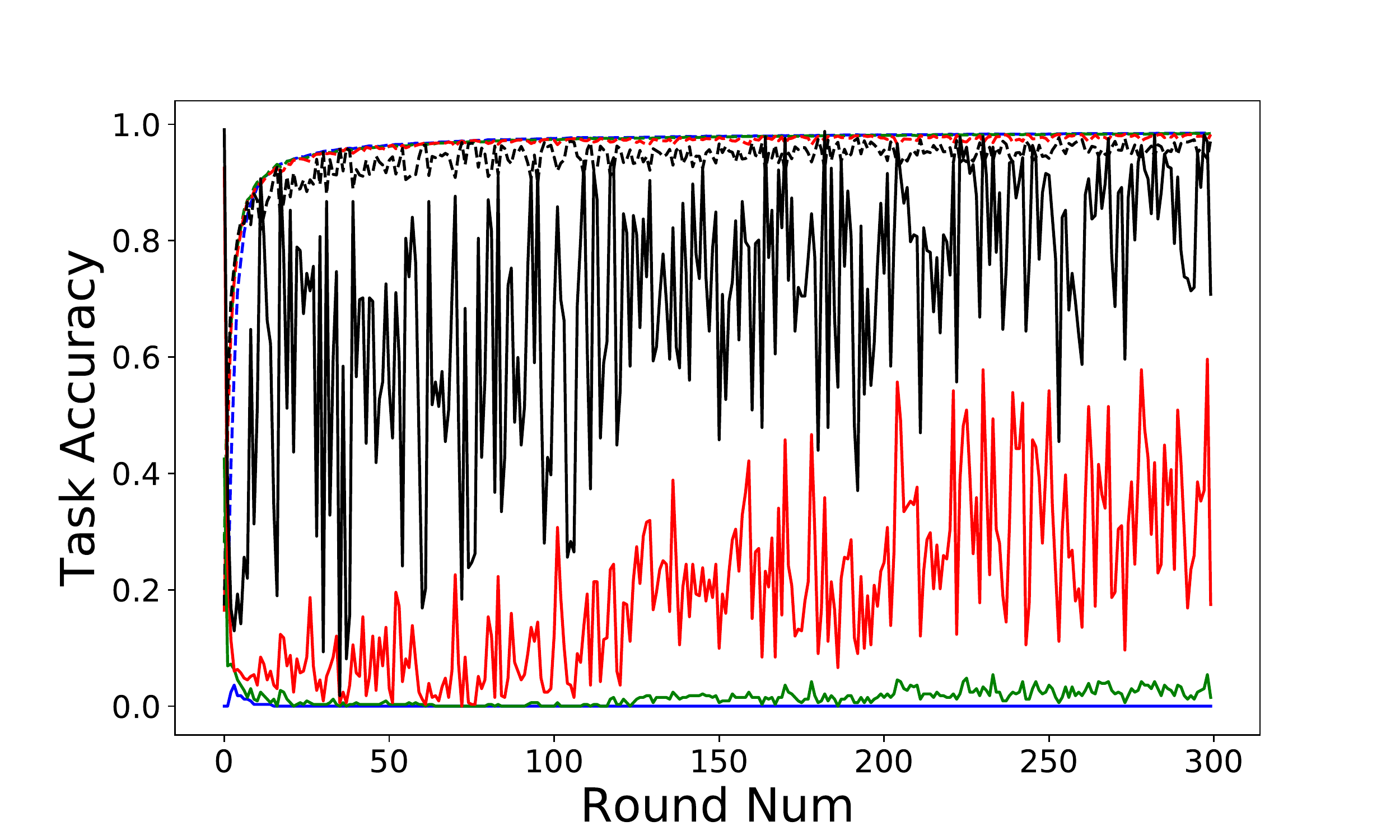}}
\subfigure[Gaussian noise (norm bound = 5). Red: $\sigma = 0$, green: $\sigma = 0.025$, blue: unattacked baseline]{%
\label{fig:noise}\includegraphics[width = 0.49\textwidth]{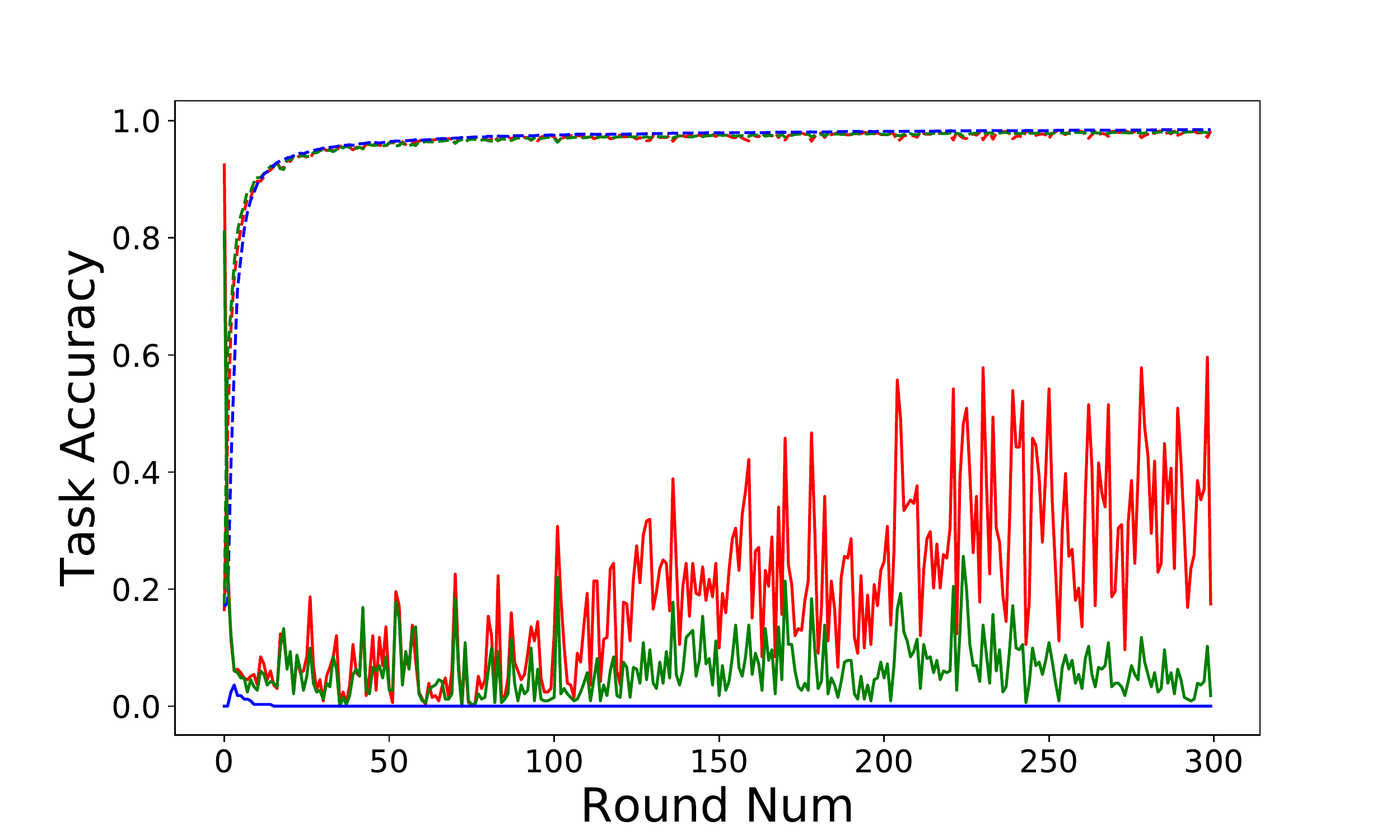}}

\caption{Effect of norm bounding and Gaussian noise. Dotted: main task. Solid: backdoor task.}

\end{figure}

\textbf{Norm bound for the update}. In Figure~\ref{fig:clip}, we consider norm bounded update from each user. We assume one attacker appears in every round, which corresponds to $\epsilon = 3.3\%$ corrupted users, and we consider norm bounds of 3, 5, and 10 (the 90 percentile of benign users' updates are below 2 for most of the rounds), which translates to $0.1, 0.17, 0.33$ norm bound for the update before boosting. We can see from the plot that selecting 3 as the norm bound will successfully mitigate the attack with almost no effect on the performance of the main task. Hence we can see that norm bounding may be a valid defense for current backdoor attacks.


\textbf{Weak differential privacy} In Figure~\ref{fig:noise}, we consider norm bounding plus adding Gaussian noise. We use norm bound of 5, which itself would not mitigate the attack, and add independent Gaussian noise with variance 0.025 to each coordinate. From the plots, we can see that adding Gaussian noise can also help mitigate the attack beyond norm clipping without hurting the overall performance much. We note that similar to previous works on differential privacy~\cite{abadi2016deep}, we do not provide a recipe for selecting the norm bound and variance of the Gaussian noise. Rather, we show that some reasonable values motivated by differential privacy literature perform well. Discovering algorithms to learn these bounds and noise values remains an interesting open research direction.


\section{Discussion}
We studied backdoor attacks and defenses for federated learning under the more realistic EMNIST dataset. In the absence of any defense, we showed that the performance of the adversary largely depends on the fraction of adversaries present. Hence, for reasonable success, there needs to be a large number of adversaries. Perhaps surprisingly norm clipping limits the success of known backdoor attacks considerably. Furthermore, adding a small amount of Gaussian noise, in addition to norm clipping, can help further mitigate the effect of adversaries. This gives rise to several interesting questions.

\textbf{Better attacks and defenses. } In the norm bounded case, multiple iterations of ``pre-boosted'' projected gradient descent may not be the best possible attack in a single round. In fact,the adversary may attempt to directly craft the ``worst-case'' model update that satisfies the norm bound (without any boosting). Moreover, if the attacker knows they can attack in multiple rounds, there might be better strategies for doing so under a norm bound. Similarly, more advanced defenses should be investigated. 

\textbf{Effect of model capacity.} Another factor that may affect the performance of backdoor attacks is the model capacity, especially that it is conjectured that backdoor attacks use the spare capacity of the deep network~\cite{liu2018fine}. How model capacity interacts with backdoor attacks is an interesting question to consider both from the theoretical and practical sides.

\textbf{Interaction of defenses with SecAgg.} Existing approaches on range proofs (e.g. BulletProof~\cite{bunz2018bulletproofs}) can guarantee this when using secure multiparty computation but how to implement them in a computationally and communication efficient way is still an active research direction.
This can also be made compatible with SecAgg if we have an efficient implementation of multi-party range proof.


\bibliographystyle{plain}
\bibliography{references}

\begin{thebibliography}{10}

\bibitem{web:TFF}
Tensorflow federated.
\newblock \url{https://www.tensorflow.org/federated}.

\bibitem{abadi2016deep}
Martin Abadi, Andy Chu, Ian Goodfellow, H~Brendan McMahan, Ilya Mironov, Kunal
  Talwar, and Li~Zhang.
\newblock Deep learning with differential privacy.
\newblock In {\em Proceedings of the 2016 ACM SIGSAC Conference on Computer and
  Communications Security}, pages 308--318. ACM, 2016.

\bibitem{bagdasaryan2018backdoor}
Eugene Bagdasaryan, Andreas Veit, Yiqing Hua, Deborah Estrin, and Vitaly
  Shmatikov.
\newblock How to backdoor federated learning, 2018.

\bibitem{pmlr-v97-bhagoji19a}
Arjun~Nitin Bhagoji, Supriyo Chakraborty, Prateek Mittal, and Seraphin Calo.
\newblock Analyzing federated learning through an adversarial lens.
\newblock In Kamalika Chaudhuri and Ruslan Salakhutdinov, editors, {\em
  Proceedings of the 36th International Conference on Machine Learning},
  volume~97 of {\em Proceedings of Machine Learning Research}, pages 634--643,
  Long Beach, California, USA, 09--15 Jun 2019. PMLR.

\bibitem{Biggio:2012:PAA:3042573.3042761}
Battista Biggio, Blaine Nelson, and Pavel Laskov.
\newblock Poisoning attacks against support vector machines.
\newblock In {\em Proceedings of the 29th International Coference on
  International Conference on Machine Learning}, ICML'12, pages 1467--1474,
  USA, 2012. Omnipress.

\bibitem{NIPS2017_6617}
Peva Blanchard, El~Mahdi El~Mhamdi, Rachid Guerraoui, and Julien Stainer.
\newblock Machine learning with adversaries: Byzantine tolerant gradient
  descent.
\newblock In I.~Guyon, U.~V. Luxburg, S.~Bengio, H.~Wallach, R.~Fergus,
  S.~Vishwanathan, and R.~Garnett, editors, {\em Advances in Neural Information
  Processing Systems 30}, pages 119--129. Curran Associates, Inc., 2017.

\bibitem{bonawitz17secagg}
Keith Bonawitz, Vladimir Ivanov, Ben Kreuter, Antonio Marcedone, H~Brendan
  McMahan, Sarvar Patel, Daniel Ramage, Aaron Segal, and Karn Seth.
\newblock Practical secure aggregation for privacy-preserving machine learning.
\newblock In {\em Proceedings of the 2017 ACM SIGSAC Conference on Computer and
  Communications Security}, pages 1175--1191. ACM, 2017.

\bibitem{bunz2018bulletproofs}
Benedikt B{\"u}nz, Jonathan Bootle, Dan Boneh, Andrew Poelstra, Pieter Wuille,
  and Greg Maxwell.
\newblock Bulletproofs: Short proofs for confidential transactions and more.
\newblock In {\em 2018 IEEE Symposium on Security and Privacy (SP)}, pages
  315--334. IEEE, 2018.

\bibitem{caldas2018leaf}
Sebastian Caldas, Peter Wu, Tian Li, Jakub Kone{\v{c}}n{\`y}, H~Brendan
  McMahan, Virginia Smith, and Ameet Talwalkar.
\newblock Leaf: A benchmark for federated settings.
\newblock {\em arXiv preprint arXiv:1812.01097}, 2018.

\bibitem{chen2017targeted}
Xinyun Chen, Chang Liu, Bo~Li, Kimberly Lu, and Dawn Song.
\newblock Targeted backdoor attacks on deep learning systems using data
  poisoning.
\newblock {\em arXiv preprint arXiv:1712.05526}, 2017.

\bibitem{cohen2017emnist}
Gregory Cohen, Saeed Afshar, Jonathan Tapson, and Andr{\'e} van Schaik.
\newblock Emnist: Extending mnist to handwritten letters.
\newblock In {\em 2017 International Joint Conference on Neural Networks
  (IJCNN)}, pages 2921--2926. IEEE, 2017.

\bibitem{damaskinos2018asynchronous}
Georgios Damaskinos, El~Mahdi~El Mhamdi, Rachid Guerraoui, Rhicheek Patra, and
  Mahsa Taziki.
\newblock Asynchronous byzantine machine learning (the case of sgd).
\newblock {\em arXiv preprint arXiv:1802.07928}, 2018.

\bibitem{pmlr-v97-diakonikolas19a}
Ilias Diakonikolas, Gautam Kamath, Daniel Kane, Jerry Li, Jacob Steinhardt, and
  Alistair Stewart.
\newblock Sever: A robust meta-algorithm for stochastic optimization.
\newblock In Kamalika Chaudhuri and Ruslan Salakhutdinov, editors, {\em
  Proceedings of the 36th International Conference on Machine Learning},
  volume~97 of {\em Proceedings of Machine Learning Research}, pages
  1596--1606, Long Beach, California, USA, 09--15 Jun 2019. PMLR.

\bibitem{Dwork06}
Cynthia Dwork, Frank Mcsherry, Kobbi Nissim, and Adam Smith.
\newblock Calibrating noise to sensitivity in private data analysis.
\newblock In {\em In Proceedings of the 3rd Theory of Cryptography Conference},
  2006.

\bibitem{pmlr-v80-mhamdi18a}
El~Mahdi El~Mhamdi, Rachid Guerraoui, and S{\'e}bastien Rouault.
\newblock The hidden vulnerability of distributed learning in {B}yzantium.
\newblock In Jennifer Dy and Andreas Krause, editors, {\em Proceedings of the
  35th International Conference on Machine Learning}, volume~80 of {\em
  Proceedings of Machine Learning Research}, pages 3521--3530,
  Stockholmsmässan, Stockholm Sweden, 10--15 Jul 2018. PMLR.

\bibitem{DBLP:journals/corr/GoodfellowSS14}
Ian~J. Goodfellow, Jonathon Shlens, and Christian Szegedy.
\newblock Explaining and harnessing adversarial examples.
\newblock In {\em 3rd International Conference on Learning Representations,
  {ICLR} 2015, San Diego, CA, USA, May 7-9, 2015, Conference Track
  Proceedings}, 2015.

\bibitem{gu2019badnets}
Tianyu Gu, Kang Liu, Brendan Dolan-Gavitt, and Siddharth Garg.
\newblock Badnets: Evaluating backdooring attacks on deep neural networks.
\newblock {\em IEEE Access}, 7:47230--47244, 2019.

\bibitem{huber1997robustness}
Peter~J Huber.
\newblock Robustness: Where are we now?
\newblock {\em Lecture Notes-Monograph Series}, pages 487--498, 1997.

\bibitem{liao2018backdoor}
Cong Liao, Haoti Zhong, Anna Squicciarini, Sencun Zhu, and David Miller.
\newblock Backdoor embedding in convolutional neural network models via
  invisible perturbation.
\newblock {\em arXiv preprint arXiv:1808.10307}, 2018.

\bibitem{liu2018fine}
Kang Liu, Brendan Dolan-Gavitt, and Siddharth Garg.
\newblock Fine-pruning: Defending against backdooring attacks on deep neural
  networks.
\newblock In {\em International Symposium on Research in Attacks, Intrusions,
  and Defenses}, pages 273--294. Springer, 2018.

\bibitem{DBLP:conf/ndss/LiuMALZW018}
Yingqi Liu, Shiqing Ma, Yousra Aafer, Wen{-}Chuan Lee, Juan Zhai, Weihang Wang,
  and Xiangyu Zhang.
\newblock Trojaning attack on neural networks.
\newblock In {\em 25th Annual Network and Distributed System Security
  Symposium, {NDSS} 2018, San Diego, California, USA, February 18-21, 2018},
  2018.

\bibitem{ma2019data}
Yuzhe Ma, Xiaojin Zhu, and Justin Hsu.
\newblock Data poisoning against differentially-private learners: Attacks and
  defenses.
\newblock {\em arXiv preprint arXiv:1903.09860}, 2019.

\bibitem{mcmahan17fedavg}
H~Brendan McMahan, Eider Moore, Daniel Ramage, Seth Hampson, and Blaise~Aguera
  y~Arcas.
\newblock Communication-efficient learning of deep networks from decentralized
  data.
\newblock In {\em Proceedings of the 20th International Conference on
  Artificial Intelligence and Statistics}, pages 1273--1282, 2017.

\bibitem{mcmahan18dplm}
H~Brendan McMahan, Daniel Ramage, Kunal Talwar, and Li~Zhang.
\newblock Learning differentially private recurrent language models.
\newblock In {\em International Conference on Learning Representations (ICLR)},
  2018.

\bibitem{Mei:2015:UMT:2886521.2886721}
Shike Mei and Xiaojin Zhu.
\newblock Using machine teaching to identify optimal training-set attacks on
  machine learners.
\newblock In {\em Proceedings of the Twenty-Ninth AAAI Conference on Artificial
  Intelligence}, AAAI'15, pages 2871--2877. AAAI Press, 2015.

\bibitem{pmlr-v97-shen19e}
Yanyao Shen and Sujay Sanghavi.
\newblock Learning with bad training data via iterative trimmed loss
  minimization.
\newblock In Kamalika Chaudhuri and Ruslan Salakhutdinov, editors, {\em
  Proceedings of the 36th International Conference on Machine Learning},
  volume~97 of {\em Proceedings of Machine Learning Research}, pages
  5739--5748, Long Beach, California, USA, 09--15 Jun 2019. PMLR.

\bibitem{DBLP:conf/sp/ShokriSSS17}
Reza Shokri, Marco Stronati, Congzheng Song, and Vitaly Shmatikov.
\newblock Membership inference attacks against machine learning models.
\newblock In {\em 2017 {IEEE} Symposium on Security and Privacy, {SP} 2017, San
  Jose, CA, USA, May 22-26, 2017}, pages 3--18, 2017.

\bibitem{steinhardt2017certified}
Jacob Steinhardt, Pang Wei~W Koh, and Percy~S Liang.
\newblock Certified defenses for data poisoning attacks.
\newblock In {\em Advances in neural information processing systems}, pages
  3517--3529, 2017.

\bibitem{DBLP:journals/corr/SzegedyZSBEGF13}
Christian Szegedy, Wojciech Zaremba, Ilya Sutskever, Joan Bruna, Dumitru Erhan,
  Ian~J. Goodfellow, and Rob Fergus.
\newblock Intriguing properties of neural networks.
\newblock In {\em 2nd International Conference on Learning Representations,
  {ICLR} 2014, Banff, AB, Canada, April 14-16, 2014, Conference Track
  Proceedings}, 2014.

\bibitem{DBLP:conf/uss/TramerZJRR16}
Florian Tram{\`{e}}r, Fan Zhang, Ari Juels, Michael~K. Reiter, and Thomas
  Ristenpart.
\newblock Stealing machine learning models via prediction apis.
\newblock In {\em 25th {USENIX} Security Symposium, {USENIX} Security 16,
  Austin, TX, USA, August 10-12, 2016.}, pages 601--618, 2016.

\bibitem{tran2018spectral}
Brandon Tran, Jerry Li, and Aleksander Madry.
\newblock Spectral signatures in backdoor attacks.
\newblock In {\em Advances in Neural Information Processing Systems}, pages
  8000--8010, 2018.

\bibitem{wang2019neural}
Bolun Wang, Yuanshun Yao, Shawn Shan, Huiying Li, Bimal Viswanath, Haitao
  Zheng, and Ben~Y Zhao.
\newblock Neural cleanse: Identifying and mitigating backdoor attacks in neural
  networks.
\newblock {\em Neural Cleanse: Identifying and Mitigating Backdoor Attacks in
  Neural Networks}, page~0, 2019.

\bibitem{Xiao:2015:SVM:2779626.2779777}
Huang Xiao, Battista Biggio, Blaine Nelson, Han Xiao, Claudia Eckert, and Fabio
  Roli.
\newblock Support vector machines under adversarial label contamination.
\newblock {\em Neurocomput.}, 160(C):53--62, July 2015.

\end{thebibliography}

\appendix
\newpage
\section{Additional figures for experiments} \label{app:figures}

\begin{figure}[tbh!]
\vspace{-25pt}
\centering

\subfigure[Attack frequency = 1/3 ($\epsilon = 1.1\%$)\label{fig:freq3}]{\includegraphics[width = 0.49\textwidth] {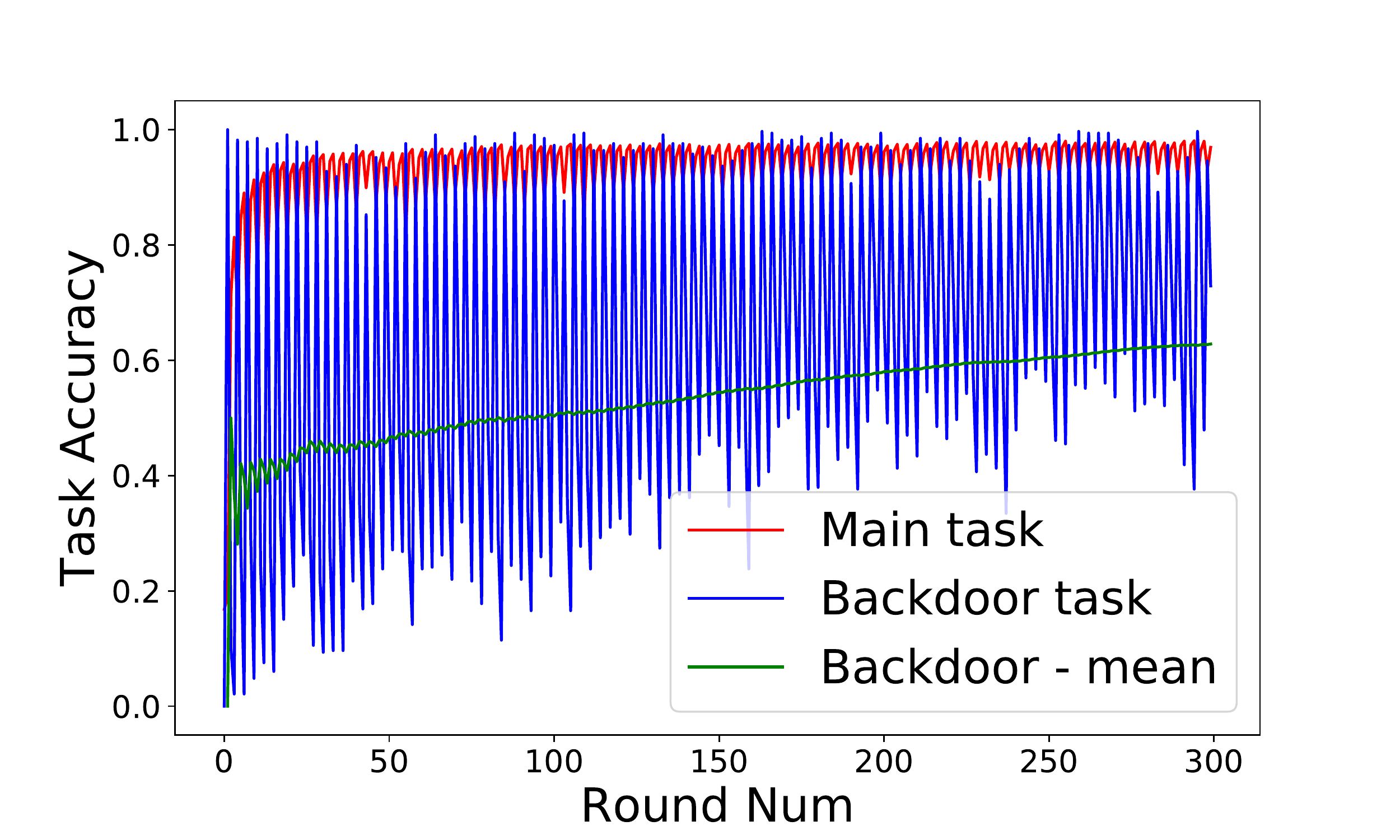}} \hfill 
\subfigure[Number of attackers = 38 ($\epsilon = 1.1\%$)\label{fig:freq3_random}]{\includegraphics[width = 0.49\textwidth]{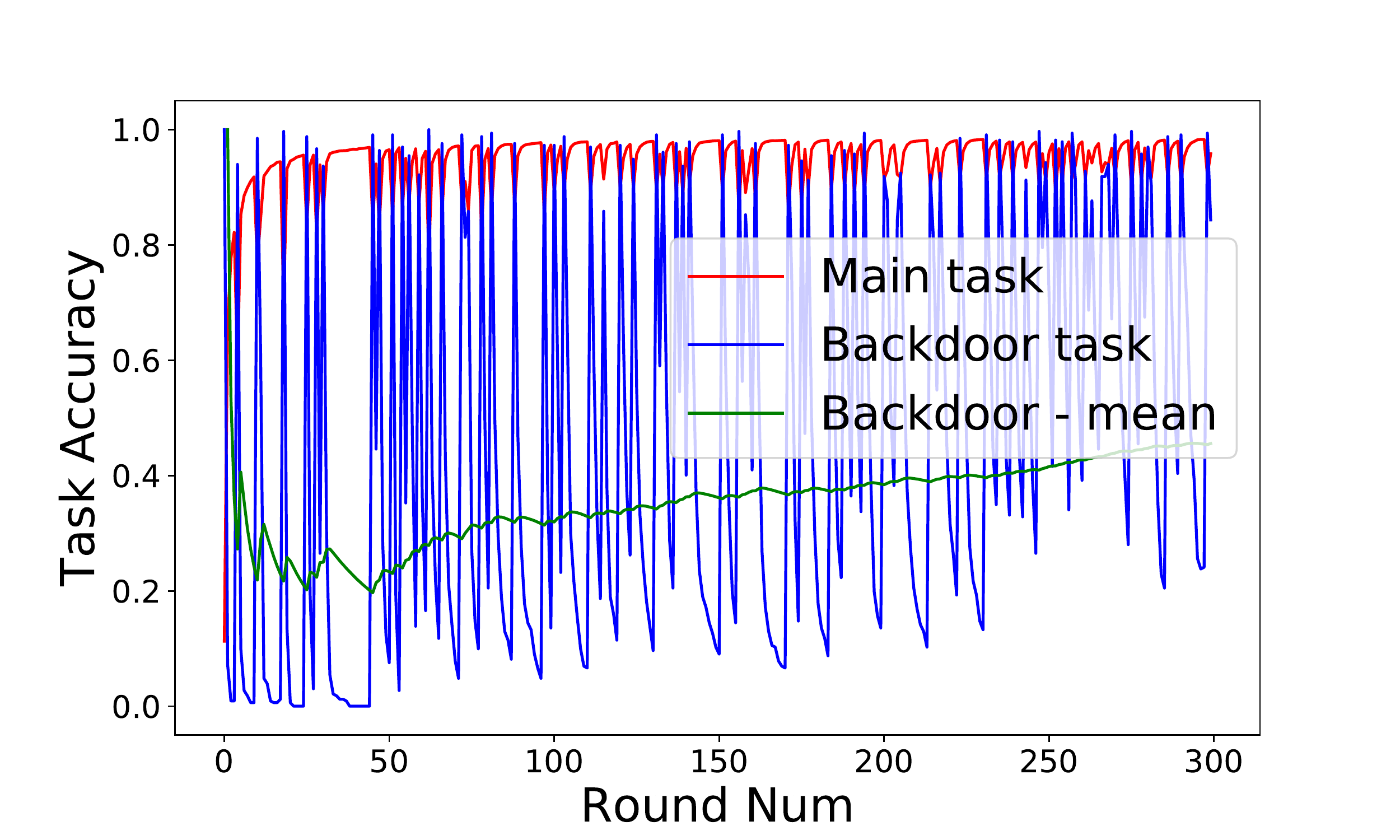}}
\subfigure[Attack frequency = 1/5 ($\epsilon = 0.67\%$)\label{fig:freq5}]{\includegraphics[width = 0.49\textwidth] {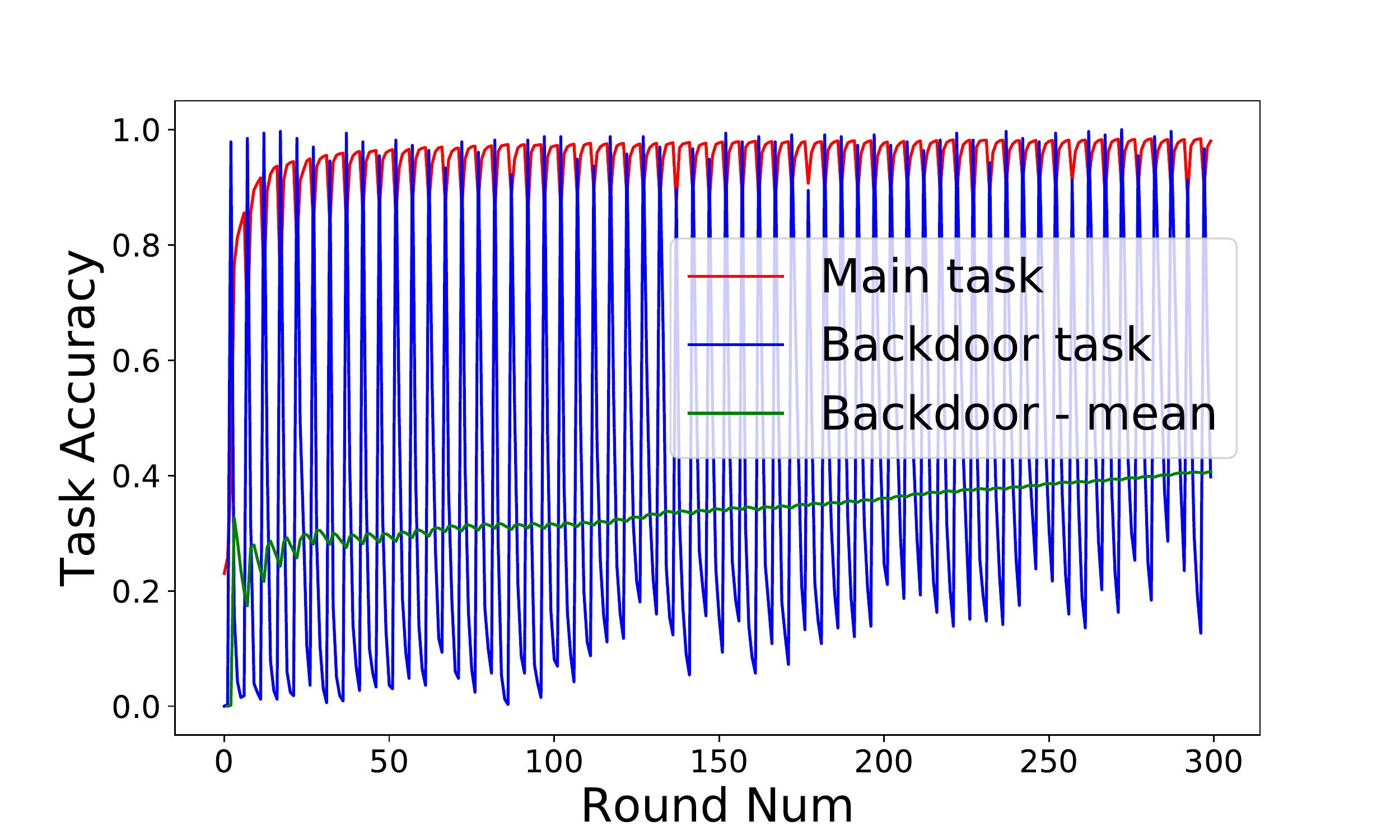}} \hfill 
\subfigure[Number of attackers = 23 ($\epsilon = 0.67\%$)\label{fig:freq5_random}]{\includegraphics[width = 0.49\textwidth]{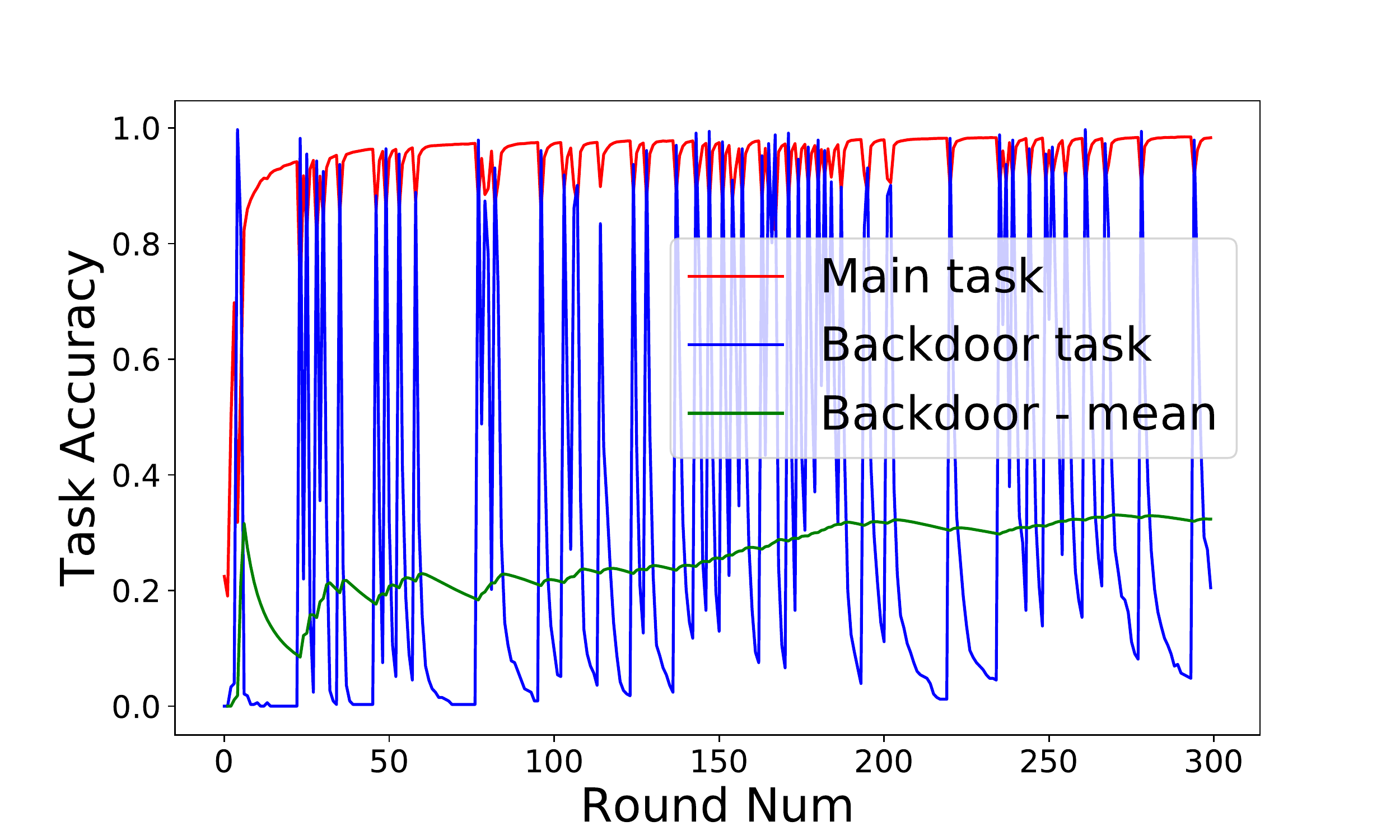}}

\caption{Unconstrained attack for fixed-frequency attacks (left column) and random sampling attack (right column) with different fractions of attackers. Green line is the cumulative mean for the backdoor accuracy.}
\vspace{-10pt}
\label{fig:unconstrained_app}
\end{figure}

\begin{figure}[tbh!]
\centering
\vspace{-10pt}
\subfigure[Attack frequency = 1/3 ($\epsilon = 1.1\%$)\label{fig:freq3_norm10}]{\includegraphics[width = 0.49\textwidth] {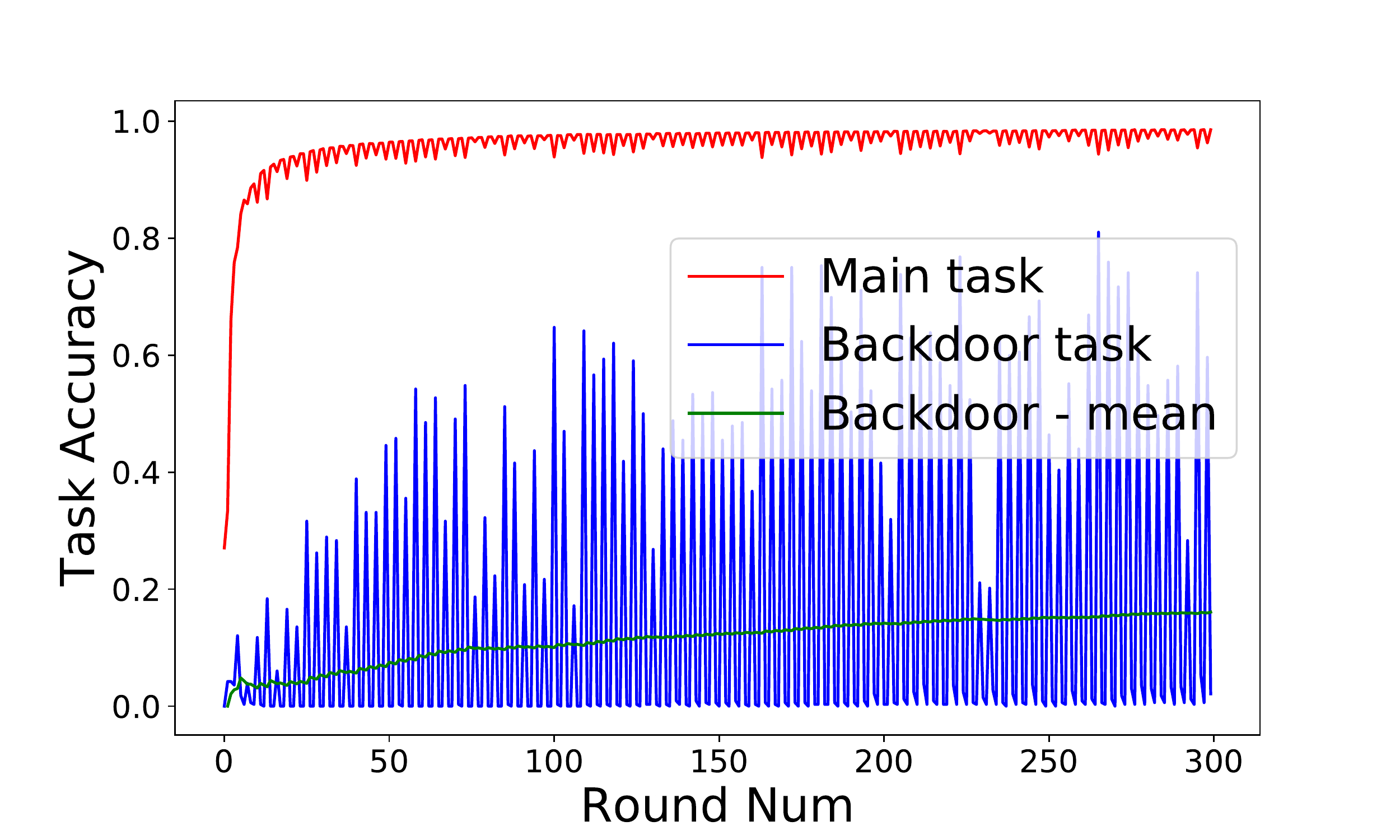}} \hfill 
\subfigure[Number of attackers = 38 ($\epsilon = 1.1\%$)\label{fig:freq3_random_norm10}]{\includegraphics[width = 0.49\textwidth]{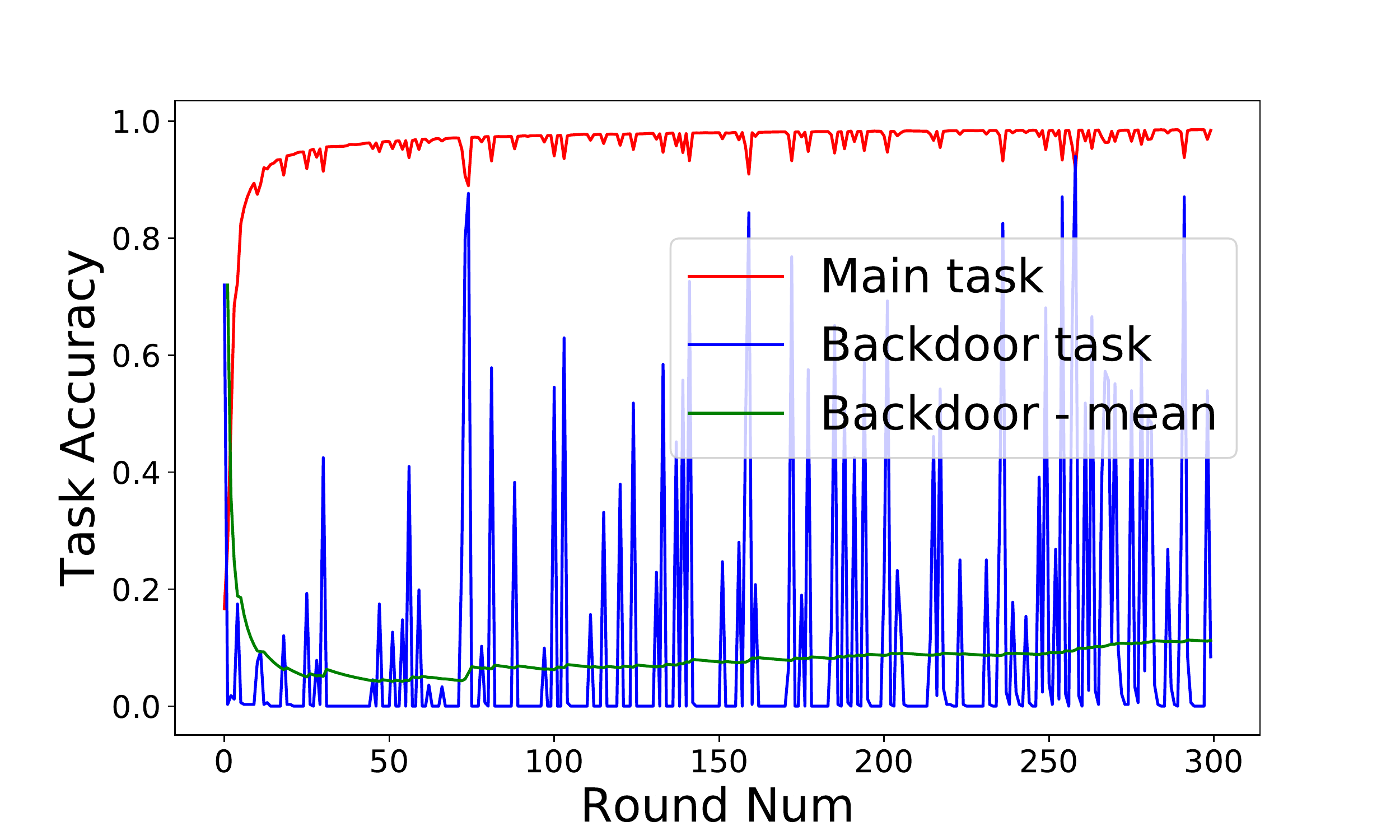}}
\subfigure[Attack frequency = 1/5 ($\epsilon = 0.67\%$)\label{fig:freq5_norm10}]{\includegraphics[width = 0.49\textwidth] {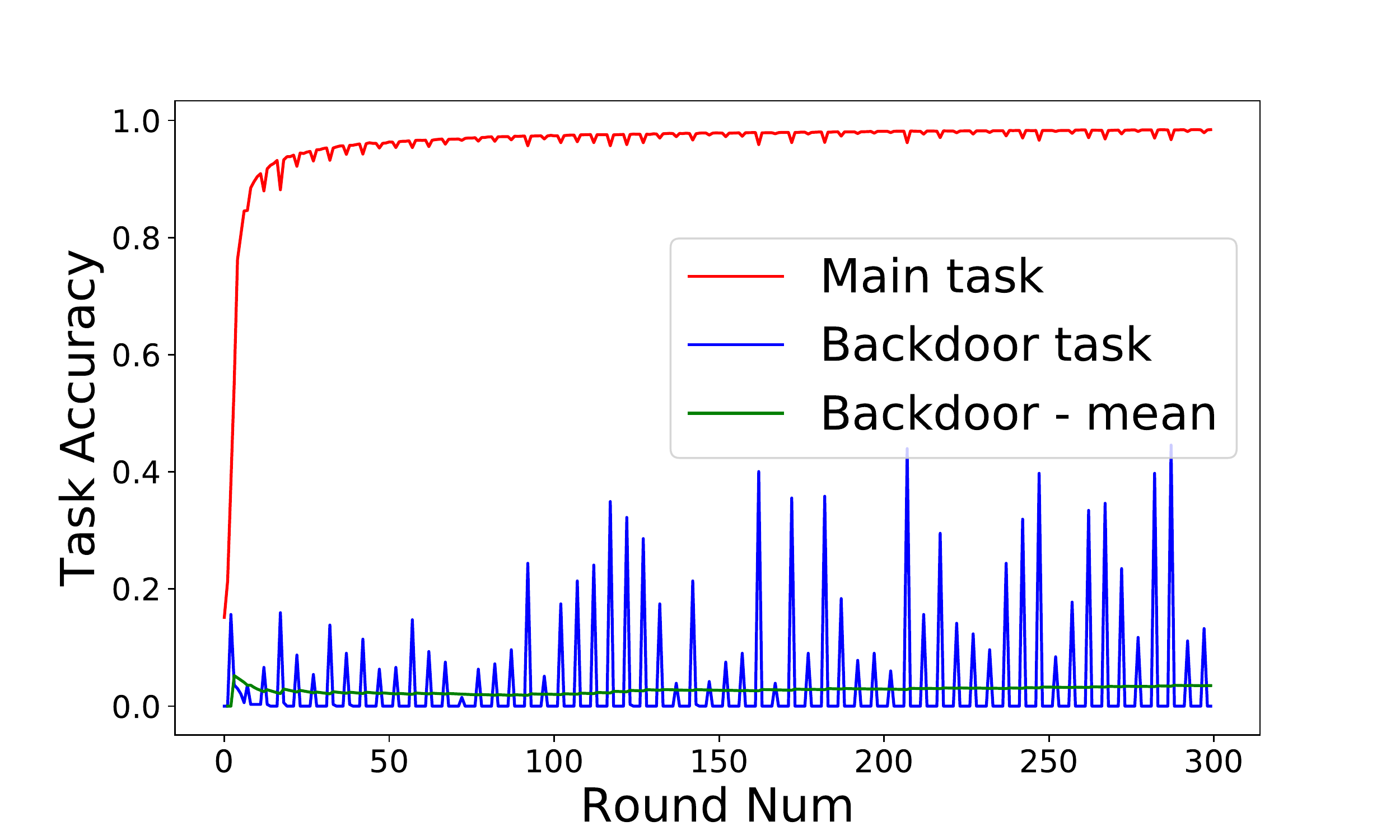}} \hfill 
\subfigure[Number of attackers = 23 ($\epsilon = 0.67\%$)\label{fig:freq5_random_norm10}]{\includegraphics[width = 0.49\textwidth]{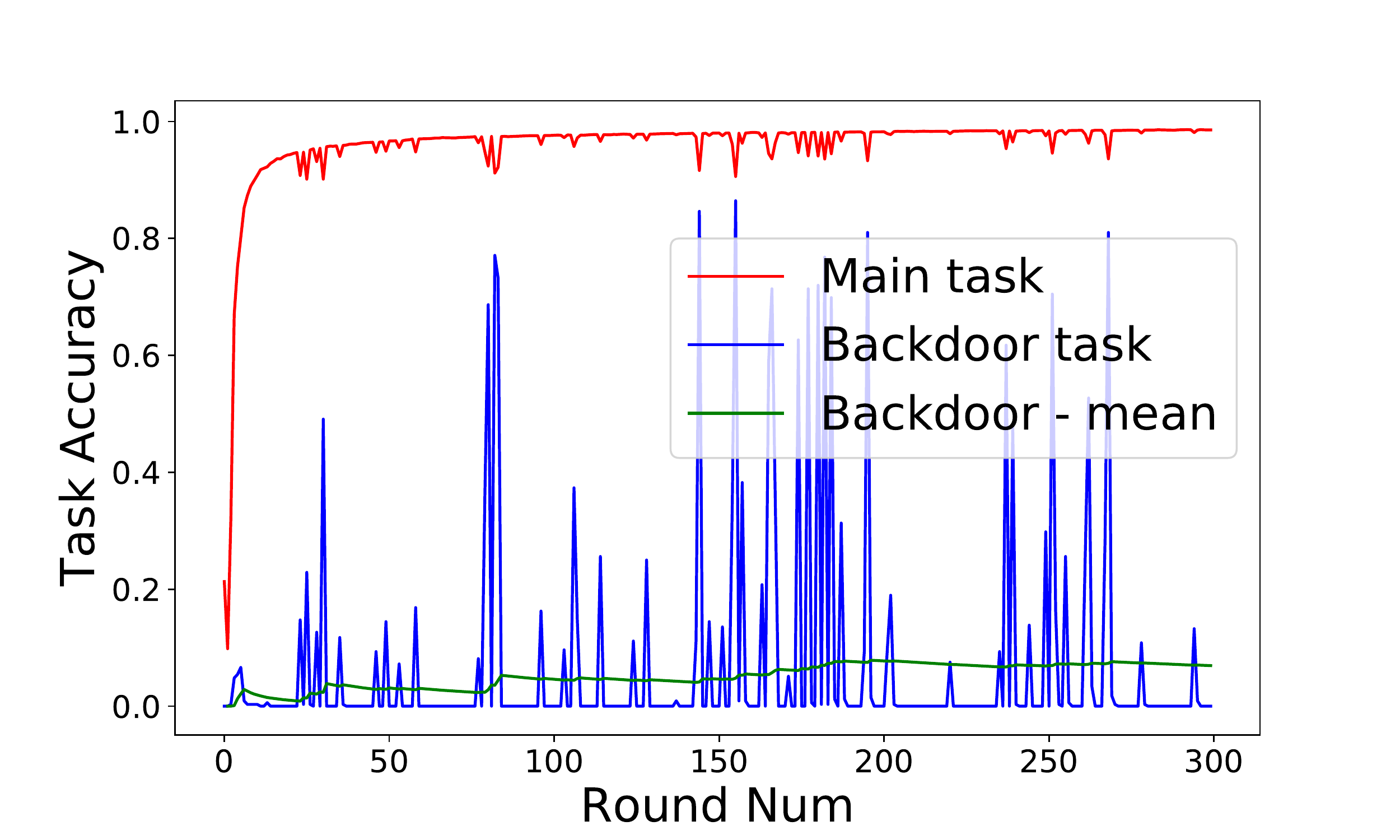}}
\caption{Constrained attack with norm bound 10 for fixed-frequency attacks (left column) and random sampling attack (right column) with different fractions of attackers. Green line is the cumulative mean for the backdoor accuracy.}
\label{fig:constrained_app}
\end{figure}

\end{document}